\documentclass[10pt,twocolumn,letterpaper]{article}

\usepackage[utf8]{inputenc}
\usepackage{graphicx}
\usepackage{subcaption}
\usepackage{multirow}
\usepackage{color}
\usepackage{xcolor}
\usepackage{amsmath}

\DeclareMathOperator*{\argmin}{arg\,min}

\usepackage{hyperref}

\newcommand{\eg}{{e.g.~}}

\title{4DHumanOutfit: a multi-subject 4D dataset of human motion sequences in varying outfits exhibiting large displacements}
\author{
  Matthieu Armando$^1$
  \and Laurence Boissieux$^2$
  \and Edmond Boyer$^2$
  \and Jean-S\'{e}bastien Franco$^2$
  \and Martin Humenberger$^1$ 
  \and Christophe Legras$^1$
  \and Vincent Leroy$^1$ 
  \and Mathieu Marsot$^2$
  \and Julien Pansiot$^2$
  \and Sergi Pujades$^2$ 
  \and Rim Rekik$^2$
  \and Gr\'egory Rogez$^1$ 
  \and Anilkumar Swamy$^{1,2}$
  \and Stefanie Wuhrer$^2$
}
\date{}

\begin{document}
\twocolumn[{
\renewcommand\twocolumn[1][]{#1}
\maketitle

\begin{center}
  \centering
  \captionsetup{type=figure}
\includegraphics[width=\linewidth]{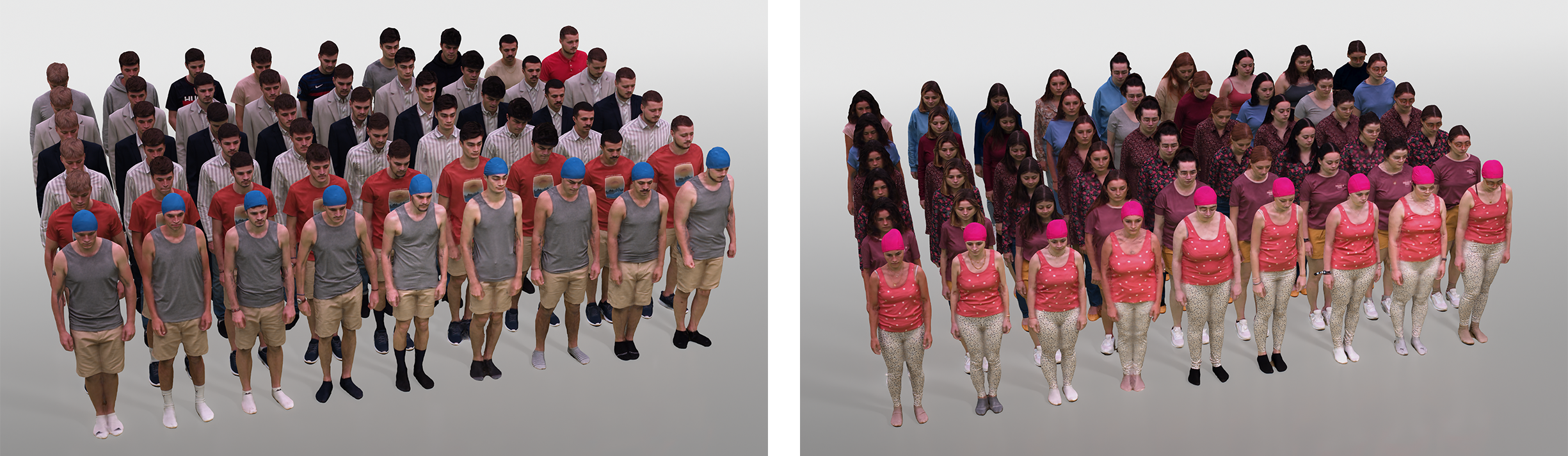}
  \captionof{figure}{Identity and outfit axes of the 4DHumanOutfit dataset. $20$ actors were captured in $7$ outfits each while performing $11$ motions per outfit. The figure shows the identities and the outfits of the subset that we release, with male actors on the left and female actors  on the right. First row shows minimal clothing, which we leverage to obtain body shape parameters by fitting a parametric body model.}
  \label{fig:teaser}
\end{center}
}]

\footnotetext[1]{NAVER LABS Europe}
\footnotetext[2]{Inria centre at the University Grenoble Alpes}
\footnotetext[3]{Authors ordered alphabetically.}

\begin{abstract}
This work presents 4DHumanOutfit, a new dataset of densely sampled spatio-temporal 4D human motion data of different actors, outfits and motions. The dataset is designed to contain different actors wearing different outfits while performing different motions in each outfit. In this way, the dataset can be seen as a cube of data containing 4D motion sequences along $3$ axes with identity, outfit and motion. This rich dataset has numerous potential applications for the processing and creation of digital humans,~\eg augmented reality, avatar creation and virtual try on. 4DHumanOutfit is released for research purposes at \url{https://kinovis.inria.fr/4dhumanoutfit/}. In addition to image data and 4D reconstructions, the dataset includes reference solutions for each axis. We present independent baselines along each axis that demonstrate the value of these reference solutions for evaluation tasks.
\end{abstract}

\section{Introduction}
4DHumanOutfit is a new dataset of 4D human motion sequences, sampled densely in space and time, with $20$ actors, dressed in $7$ outfits each, and performing $11$ motions exhibiting large displacements in each outfit. We designed 4DHumanOutfit to enable the combined analysis of shape, outfit and motions with humans. This results in a dataset shaped as a cube of data containing 4D motion sequences with three different factors that vary along the axes identity, outfit, and motion. Fig.~\ref{fig:teaser} illustrates the morphology and clothing axes of this cube.

Analyzing and modeling the dynamics of human garments during motion and across actors is a well-studied problem in computer vision and computer graphics, with the goals of understanding human motion from partial data and generating realistic digital human animations. This has applications in video understanding, including action and fashion recognition; telepresence, including virtual change rooms and fashion transfers; and entertainment, including animation content generation. Many existing works study this problem from a data-driven perspective, where the goal is to learn motion dynamics from example data. To facilitate these studies, three main types of datasets of humans in motion have been introduced. The first type of dataset contains 2D videos of dressed humans in motion~\eg\cite{wang_21}, which allows to capture the appearance of rich clothing dynamics observed in real garments. More recently, large-scale 4D datasets of minimally clothed 3D human bodies in motion have been published, either captured using acquisition platforms~\eg\cite{Dyna:SIGGRAPH:2015} or computed by fitting models to sparse motion capture data~\eg\cite{mahmood2019amass},
which allow to learn detailed 3D body shape deformations over time. To enhance such datasets with garments, recent works use physical simulation to drape clothing on this 4D data~\eg\cite{santesteban_22},  enabling therefore to model clothing dynamics for synthetically generated garments. 4DHumanOutfit  contributes 4D data sampled densely in space and time of human bodies captured in different outfits and different motions. This combines the advantages of existing 2D datasets of capturing dynamic behaviour of layered clothing, including complex dynamics caused by seams and friction, with the advantages of existing 4D datasets of containing 3D shape information, including fine-scale geometric details.

The data we present has been captured in a multi-view acquisition platform that acquires $68$ synchronized RGB streams at $50$ frames per second, which are subsequently used to reconstruct densely sampled 3D geometric models with texture information per frame. For this dataset, we provide the RGB videos with masked background, and the reconstructed motion sequences in 4D in different spatial resolutions.

To demonstrate the potential of our dataset, we perform a baseline evaluation along each of the three axis independently. To this end, we introduce three tasks together with evaluation protocols. For the identity axis, we aim to predict the body shape of an identity given a 4D motion sequence of an actor wearing an arbitrary outfit. As reference solution for evaluating this task, we provide sequences captured in minimal clothing with body shapes resulting from fitting a standard parametric human body model~\cite{loper2015smpl} to the data. For the outfit axis, we aim to retrieve the outfit in a standardized pose from a given 4D motion sequence of an actor wearing an arbitrary outfit. As a reference solution to evaluate this task, we provide static scans of the outfit acquired on a mannequin. For the motion axis, we aim to retarget motion between actors from a 4D motion sequence showing the source motion to a static 3D scan of the target actor. When applied to data within the datacube, reference solutions are available as every actor was acquired performing each of the motions in each of the outfits. These tasks demonstrate that each axis of the datacube provides unique information that can be exploited in a large variety of practical applications.

The main contributions of this work are:
\begin{itemize}
    \item The introduction of 4DHumanOutfit, a datacube of dynamic 4D human motion of $20$ actors in $7$ outfits each, performing $11$ motions in each outfit, i.e.~1540 sequences in total. A large subset of $18$ actors in $6$ outfits and $10$ motions is released for research purposes.
    \item The proposition of associated evaluation protocols and reference solutions for three tasks, along the identity, outfit, and motion axes of the datacube.
\end{itemize}

\section{Related Work}

Capturing humans in clothing has attracted many efforts in computer vision and graphics. Existing works can be mainly clustered into datasets containing
(i) 2D images of clothed persons, (ii) synthetic 3D models, and (iii) 3D scans of humans in motion, different identities and various outfits.
We review them briefly to highlight how 4DHumanOutfit goes beyond the state of the art. 

\subsection{2D fashion datasets}
Many works focus on the creation of datasets of 2D images, as these are relatively easy to acquire. This includes early works, such as the Fashionista dataset~\cite{yamaguchi2012parsing} or works targeting to describe clothing with semantic attributes~\cite{chen2012describing},
as well as more recent datasets, such as FashionPedia~\cite{jia2020fashionpedia}, among many others made available for research purposes.
A recent survey~\cite{cheng2021fashion} provides an exhaustive list of published datasets until $2020$, with a comprehensive classification of the tasks that can be addressed with 2D data. These include landmark detection, clothing parsing, retrieval, and attribute recognition. In addition, these datasets have allowed to tackle the task of virtual try on, where one can create high quality, compelling images~\cite{liu2016deepfashion, han2018viton} or videos~\cite{dong2019fw} of how a person would look like wearing a given clothing in a target pose. While the generated images reach impressive realistic quality, their applicability is yet limited, as they cannot be used for actual metric fit assessment.

\subsection{Synthetic datasets}
Capturing data with cameras or 3D scanners usually requires tremendous human and material efforts. In order to circumvent this issue, generated synthetic datasets of clothed humans can be considered.  In this category works have proposed different datasets, with one or multiple characters in different settings, by leveraging 3D editing tools, such as MakeHuman~\cite{makehuman}, Mixamo~\cite{mixamo}, or human body models, such as SMPL~\cite{loper2015smpl}.
For example, several datasets such as SURREAL~\cite{varol2017learning}, MHOF~\cite{ranjan2020learning} or LTSH\cite{hoffmann2019learning}, place 3D models of humans on background images.
These datasets have been designed to address the task of 2D or 3D pose estimation from a single image.
Other datsets, such as AGORA~\cite{patel2021agora}, have increased the challenge by 
including images of multiple dressed persons with plausible interactions with the environment.
As all these images are static and lack 3D realism, they do not capture and model the complexity of real cloths' dynamics.

To include dynamics in the data, most works leverage physics simulators, which can account for the type of clothing through physical parameters. 
Since the early work of Guan et al.~\cite{guan2012drape}, several methods have considered different clothing parameters, which allow creating plausible variations in the wrinkle patterns present on cloths.
Synthetic datasets, with modest sizes such as BCNet~\cite{jiang2020bcnet}, or larger scale datasets, such as 3DPeople~\cite{pumarola20193dpeople} and Cloth3D~\cite{bertiche2020cloth3d, madadi2020learning} provide 3D models of clothed humans. The last two explicitly explore the three dimensions of identity, motion, and clothing. While the explored range of subjects, poses, and cloth variations is impressive, the realism of these data are limited by the accuracy of the simulator used to create the data. The proposed 4DHumanOutfit dataset takes an alternative approach by capturing reality in a multi-view studio.

Interesting recent works have even modeled synthetic clothing at the sewing pattern level, allowing to automatically adjust the garment size to a personalized shape~\cite{korosteleva2021generating}. In our 4DHumanOutfit dataset we also release scans of the clothes on mannequins, which could allow to work on the clothes at the {\it sewing pattern level}.

All these works providing synthetic datasets have highly contributed to the community to advance the algorithmic approaches.
With our work we argue that the acquisition of actual humans performing dynamic motion in varied clothing is necessary to validate the applicability of existing approaches to real data.

\subsection{Scanned 3D humans datasets}

With 4DHumanOutfit we explore the three axis of motion, identity, and clothing. We briefly review existing datasets that consider similar axes. 

\paragraph{Motion.}
Human pose plays a crucial role in many application fields, such as medicine, sports or graphics, thus it has attracted many research efforts.

{\bf Marker based motion capture.}
A classical approach to capture human motion is to use motion capture (MoCap) systems with \eg reflective markers. 
Following the pioneering dataset HumanEva~\cite{sigal2010humaneva}, many other datasets have been acquired:  Human3.6M~\cite{ionescu2013human3}, Total Capture~\cite{trumble2017total}, AMASS~\cite{mahmood2019amass} or HUMAN4D~\cite{chatzitofis2020human4d}. 
The reflective markers allow to extract a good approximation of the pose, which is considered ground truth. 
Other modalities, such as video, depth sensors or inertial sensors, are simultaneously acquired.
From this paired data, researchers have studied how to infer a pose from these other modalities. 
In addition, massive datasets, such as AMASS~\cite{mahmood2019amass}, have allowed to learn human pose priors which are widely used in the literature. While yielding precise information on poses with the marker locations, MoCap systems provide only sparse information on motion and imply complex setups with markers to be placed on subjects. 

{\bf Markerless approaches.} Another strategy to capture the human pose and motion is to use markerless systems,  relying for that purpose on  monocular settings~\cite{mehta2017monocular, alldieck2018video, xu2018monoperfcap, mehta2018single, wang_21}; on  passive multi-view video systems, like for instance HUMBI~\cite{yoon2021humbi} and
Hi4D~\cite{yin2023hi4d}, the PanopticStudio~\cite{Simon_2017_CVPR, Joo_2017_TPAMI}; or on active systems such as 3DMD, used to acquire for example DynamicFaust~\cite{dfaust:CVPR:2017}, Flame~\cite{flame2017} or Mano~\cite{mano2017}. Depth camera setups have also been used to capture 4D motion sequences of multiple subjects~\cite{cai2022humman}.
For our work we use the Kinovis acquisition platform~\cite{kinovis}, a passive multi-camera system with a wide acquisition volume that enables dynamic displacements of the subjects and rich clothing dynamics. 

\paragraph{Identity.}
In the identity axis, the seminal CAESAR dataset~\cite{Robinette1999}, created to study body morphology and clothing sizing purposes, contains scans of over $4500$ individuals in $3$ static poses each. Further datasets have scanned different persons in static~\cite{Anguelov2005,Hasler2009a, Bogo:CVPR:2014} or dynamic~\cite{Dyna:SIGGRAPH:2015} situations.
Hasler et al.~\cite{Hasler2009a} provide a total of $500$ static poses of $114$ individuals, while the FAUST dataset~\cite{Bogo:CVPR:2014} contains $10$ individuals in $10$ static poses each.
DYNA~\cite{Dyna:SIGGRAPH:2015} contains $10$ individuals in a total of $129$ sequences, which have been accurately registered for benchmarking in the DynamicFAUST dataset~\cite{dfaust:CVPR:2017}.
All these datasets have allowed the study of identity and static or dynamic pose, but have not considered the clothing axis.

\paragraph{Outfit.}
Early efforts have focused on capturing and analyzing sequences of few individuals captured in a single outfit~\eg\cite{starck_2007,Aguiar2008,Vlasic2009}.
More recently, different tasks related to clothing have motivated the creation of additional 3D datasets of real clothed humans.

For example, to explore how different sizes of cloths drape on the same human, the dataset SIZER~\cite{tiwari2020sizer} captured around $2000$ static scans, from $100$ subjects wearing clothes of different sizes. As all subjects strike the same A-pose, this dataset does not allow to study cloth dynamics.

To tackle the problem of estimating the shape under clothing, Yang et al.~\cite{Yang_2016} and Zhang et al.~\cite{Zhang_2017_CVPR} acquired scans of different subjects, with and without clothing, performing several motions.
These datasets consider $6$ subjects, $3$ motions, and $3$ outfits for Yang et al.~\cite{Yang_2016} and $5$ subjects, $3$, motions and $2$ outfits where 4DHumanOutfit considers $20$ subjects in $7$ outfits and $11$ motions.
Other datasets have been acquired with  consumer RGBD sensors~\cite{zheng2019deephuman}, providing lower quality than 4DHumanOutfit.

To learn a generative model of clothing, the dataset CAPE~\cite{Ma_2020_CVPR} was released. It also includes scans and SMPL mesh registrations from the ClothCap work~\cite{pons2017clothcap} and contains $15$ subjects in $4$ different  outfits performing different motions. The 4DHumanOutfit dataset is larger with $20$ subjects and $7$ clothing styles with richer dynamics. 
In addition, the systematic acquisition of all subjects performing very similar motions in all outfits provides an unprecedented opportunity to study how dynamic cloth deformations behave depending on identity, motion, and clothing.

\section{4DHumanOutfit Dataset}

In the following we detail the acquisition setup, the constitution of a cohort of subjects, the selected outfits, motions, and their captures.

\subsection{Data acquisition}

\begin{figure}[h!]
  \centering
    \includegraphics[width=\linewidth]{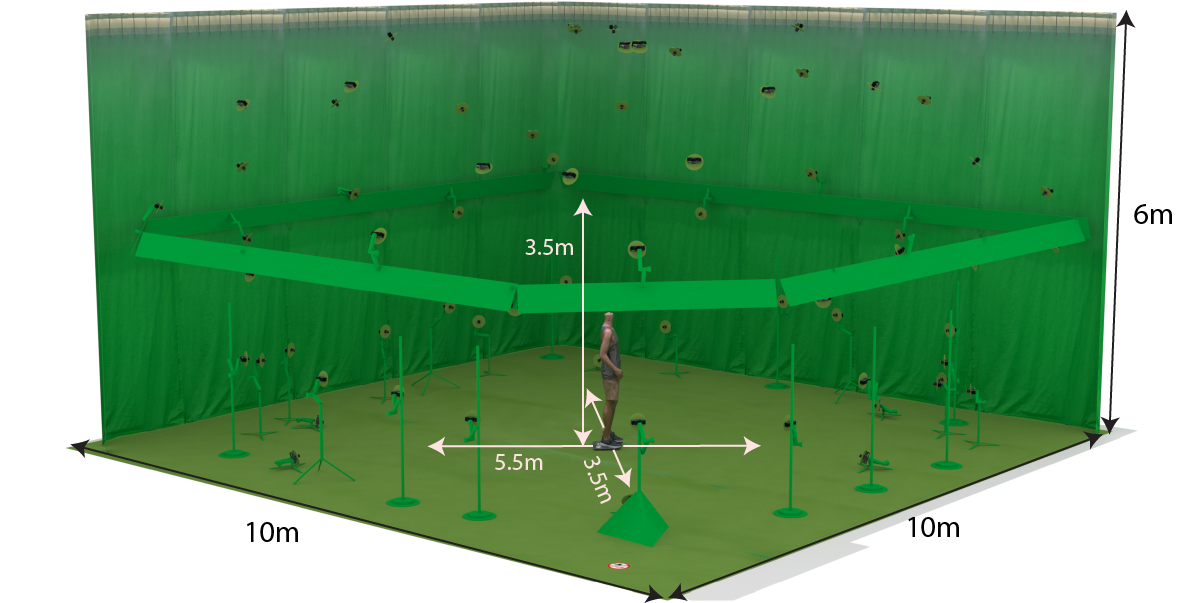}
  \caption{Multi-cameras acquisition platform~\cite{kinovis} used to capture the 4DHumanOutfit dataset.}
  \label{fig:platform}
\end{figure}

Motion sequences were captured by $68$ calibrated RGB cameras ($4$ megapixels, $50$ frames per second, focal lengths between $8$ and $16 mm$) positioned roughly on a half-ellipsoid with radii $4m$ and $5m$ and height $5m$ looking towards the stage centre, for an average image resolution of $2.5mm$ per pixel at the scene centre. The total capture area covers a length of $5.5m$ and a width of $3.5m$. Fig.~\ref{fig:platform} shows the multi-camera platform~\cite{kinovis}.

\begin{figure*}[h!]
  \centering
    \includegraphics[width=\linewidth]{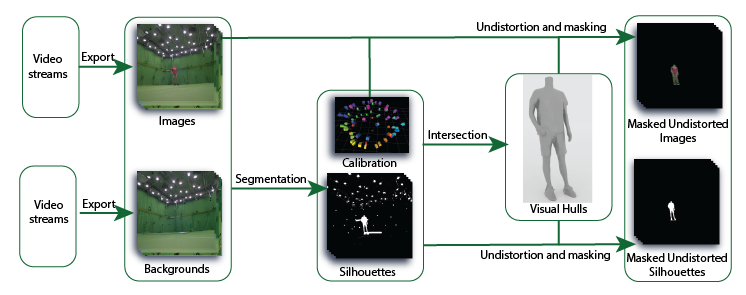}
  \caption{Data capture pipeline. Acquisition, production of visual hulls~\cite{4dviews}, and pre-processing of input images.}
  \label{fig:pipeline1}
\end{figure*}

\begin{figure*}[h!]
  \centering
    \includegraphics[width=\linewidth]{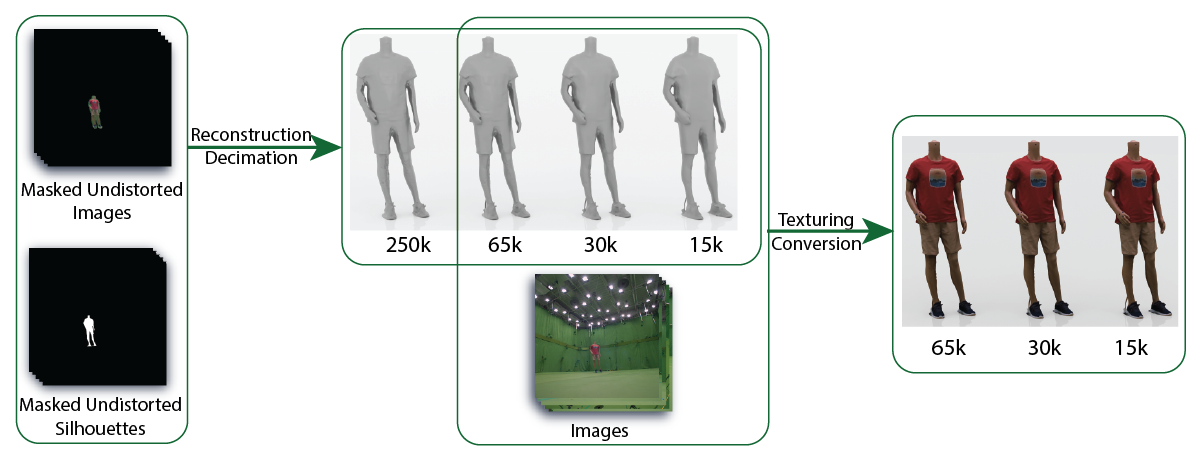}
  \caption{3D reconstruction pipeline. Multi-view reconstruction~\cite{leroy:hal-01567758} and final textured meshes. }
  \label{fig:pipeline2}
\end{figure*}

The capture and reconstruction pipelines, shown in Fig.~\ref{fig:pipeline1} and~\ref{fig:pipeline2}, consist of several steps. First, synchronized video streams are acquired and silhouettes are segmented with the software~\cite{4dviews} of the multi-camera platform~\cite{kinovis}. Second, the resulting images and silhouettes are undistorted using the calibration information and masked with projections of inflated visual hulls. This significantly reduces the size of the data. 

3D reconstructions are computed independently per frame, which results in a densely sampled 3D mesh per time instant. These meshes are obtained by performing multi-view reconstruction~\cite{leroy:hal-01567758} on the undistorted and masked images.
We decimate the resulting reconstruction into lower resolutions of 250k, 65k, 30k, and 15k vertices. The 3 lower resolution meshes are texture mapped using the capture platform software~\cite{4dviews}, which is not designed to handle higher resolutions. 

\subsection{Identities}

\begin{figure}[h!]
  \centering
    \includegraphics[width=\linewidth]{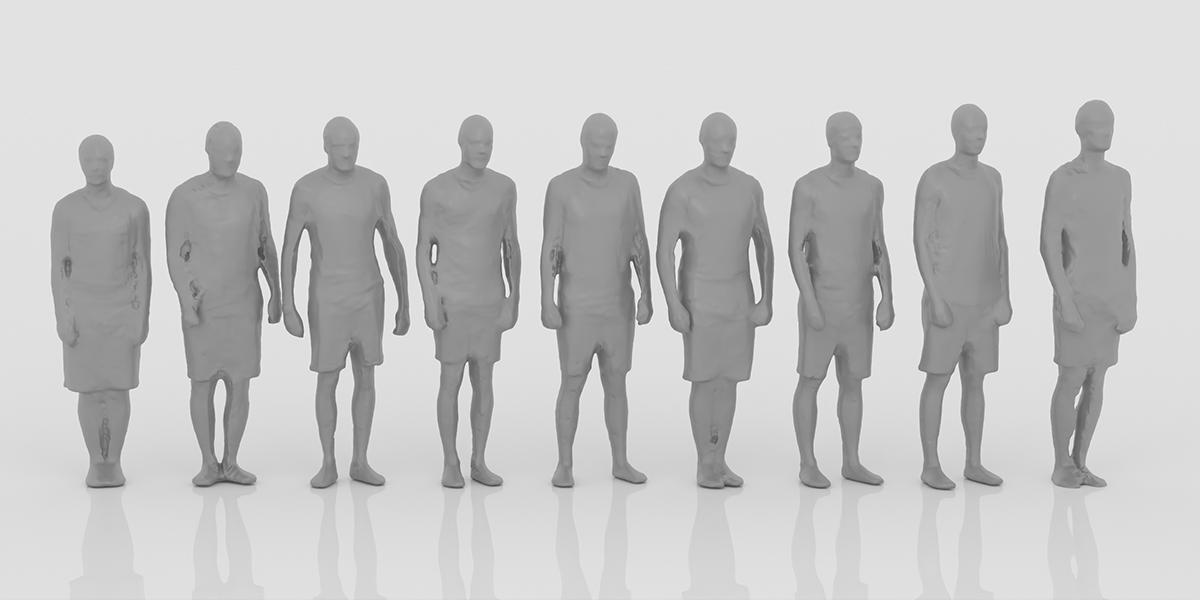}
    \includegraphics[width=\linewidth]{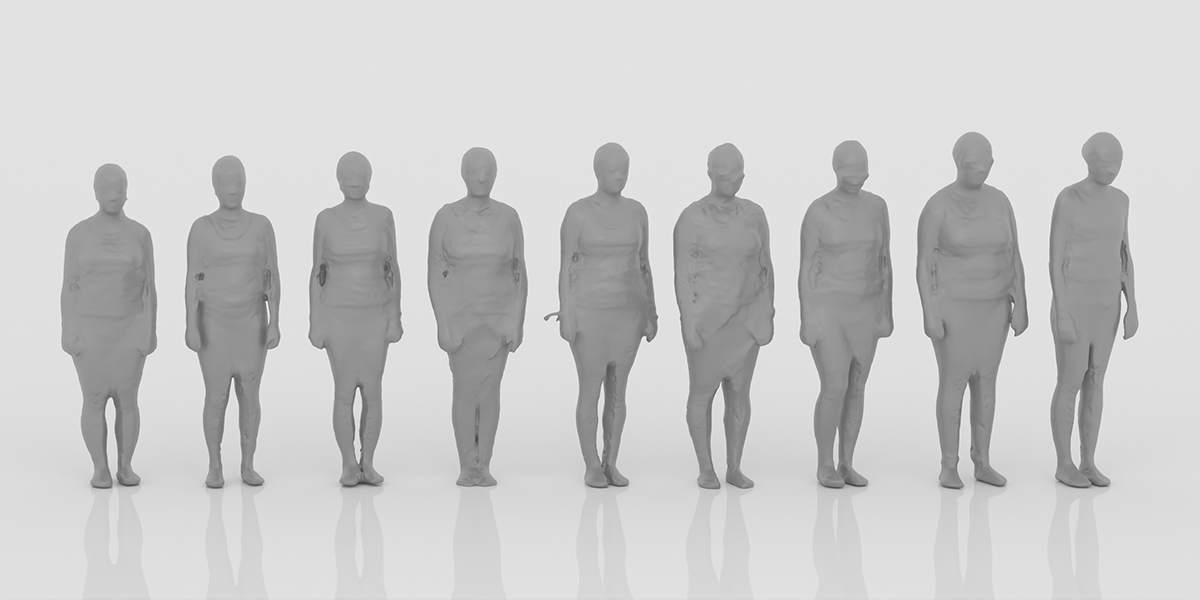}
  \caption{Morphologies. Men (top), women (bottom).}
  \label{fig:morphologies}
\end{figure}

$10$ females and $10$ males were recorded. 
The participants were empirically chosen to cover the main variations of body shape according to eigenvectors computed on the CAESAR dataset~\cite{Robinette1999}. The body shapes of all actors are shown in Fig.~\ref{fig:morphologies}. 

We release data of $9$ female and $9$ male actors, and keep the remaining data hidden to allow for future evaluations on unseen data. 

\subsection{Outfits}

\begin{figure}[h!]
  \centering
  \includegraphics[width=\linewidth]{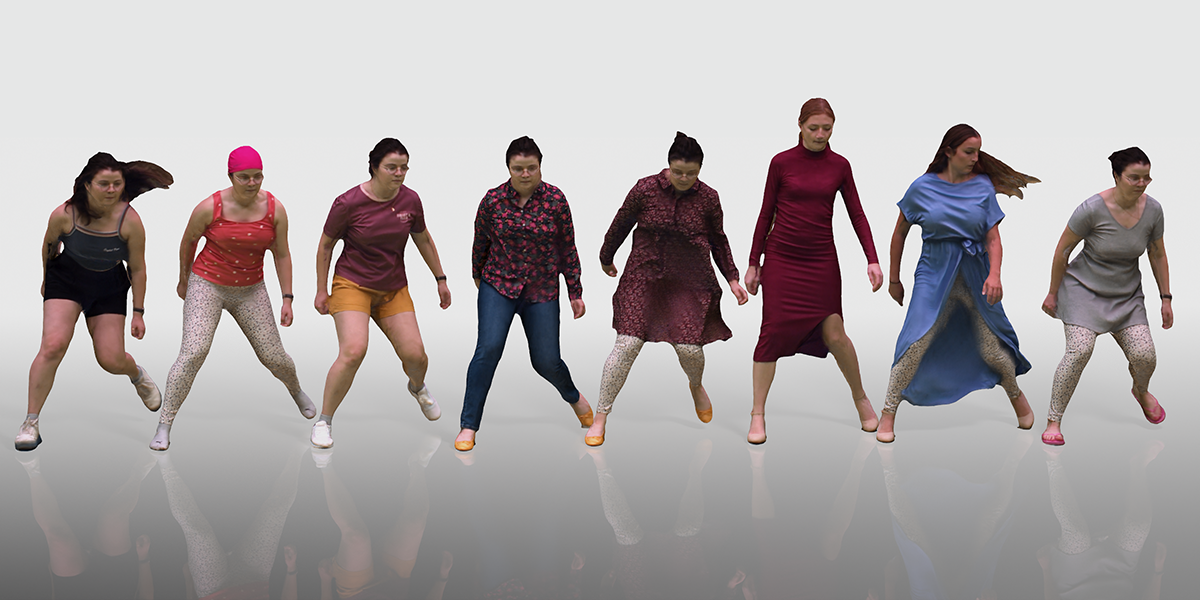}
  \caption{Female outfits. From left to right: own, tig, sho, jea, cos, opt1, opt2, opt3.}
  \label{fig:wcloth}
\end{figure}

\paragraph{Motion recordings}
Each actor was recorded wearing their own arbitrary clothes and $6$ additional outfits, chosen to cover a wide range of typical casual European clothing. The outfits differ in terms of their fit, from tight to  wide, and are made of various materials, which results in rich dynamic behaviour during motion. Outfits are different for males and females.
The following outfits, shown in Fig.~\ref{fig:wcloth}, were used for women:
\begin{itemize}
    \item\textbf{own} the actor's own clothes, unique to each actor, with the purpose to increase variability;
    \item\textbf{tig} socks, dotted white leggings, dotted salmon tank top, pink swimming cap (minimal clothing);
    \item\textbf{sho} white and pink sneakers, yellow shorts, purple T-shirt;
    \item\textbf{jea} yellow ballerinas, jeans, green and pink flowery shirt;
    \item\textbf{cos} yellow ballerinas, jeans, flowery purple dress;
    \item\textbf{opt} pink flip-flops or cream high heels, short grey dress or long loose blue dress or long tight red dress; as apparel for females offers more diversity than for males, we chose 3 optional outfits, using different materials and shapes to increase variability.
    \item\textbf{hidden} we recorded one additional outfit which will not be released to allow for future evaluations on unobserved data.
    \end{itemize}

\begin{figure}[h!]
  \centering
  \includegraphics[width=\linewidth]{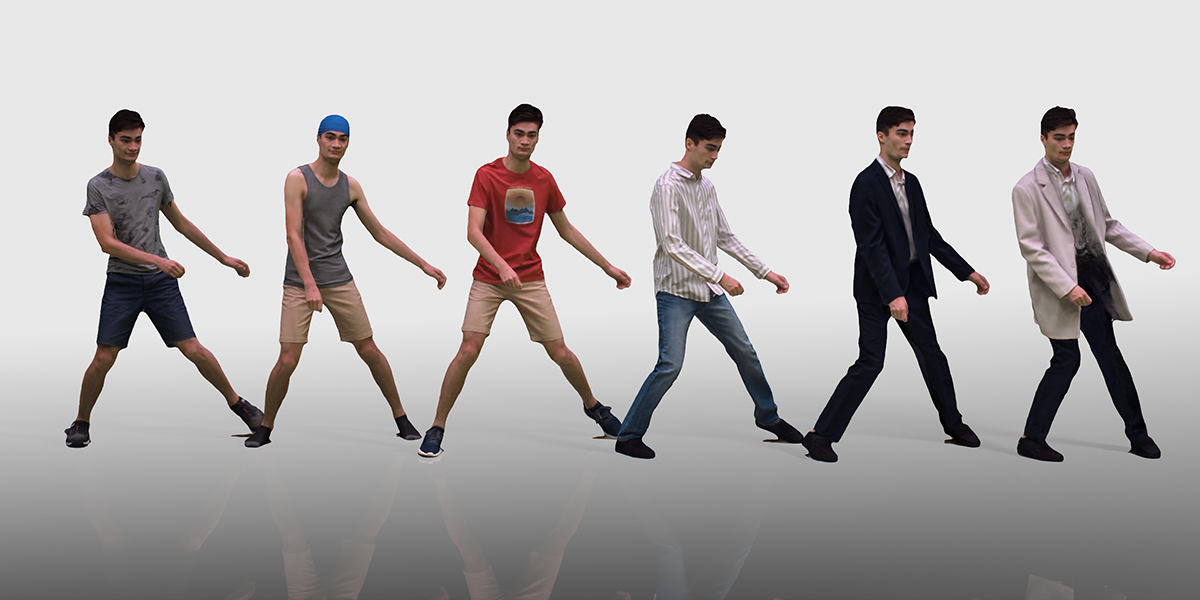}
  \caption{Male outfits. From left to right: own, tig, sho, jea, cos, opt.}
  \label{fig:mcloth}
\end{figure}

For men, the following outfits, shown in Fig.~\ref{fig:mcloth}, were used:
\begin{itemize}
    \item\textbf{own} the actor's own clothes;
    \item\textbf{tig} socks, beige shorts, grey tank top, blue swimming cap (minimal clothing);
    \item\textbf{sho} blue and white sneakers, beige shorts, orange T-shirt with picture;
    \item\textbf{jea} black moccasins, jeans, grey and white striped shirt;
    \item\textbf{cos} black moccasins, dark costume trousers, grey and white striped shirt, dark costume jacket;
    \item\textbf{opt} black moccasins, dark costume trousers, grey and white striped shirt, beige trench coat;
    \item\textbf{hidden} we recorded one additional outfit which will not be released to allow for future evaluations on unobserved data.
\end{itemize}

Each actor was recorded in $7$ outfits, including all non-optional ones and one optional outfit.

\paragraph{Reference scans}

In addition to clothed human motion data, we acquired scans of each outfit. They can serve as reference solution for an outfit retrieval task. These models were acquired using two different systems as static scans of each outfit worn by a standard mannequin. The first scanning system used is our multi-camera platform; it was used to scan the mannequins without clothing and to record $8$ scans for each outfit to allow for some natural variability in the clothing folds. The second scanning system is an Artex Eva structured light scanner, with scan resolution of about 1500k vertices.

\begin{figure}[h!]
  \centering
    \includegraphics[width=0.5\linewidth]{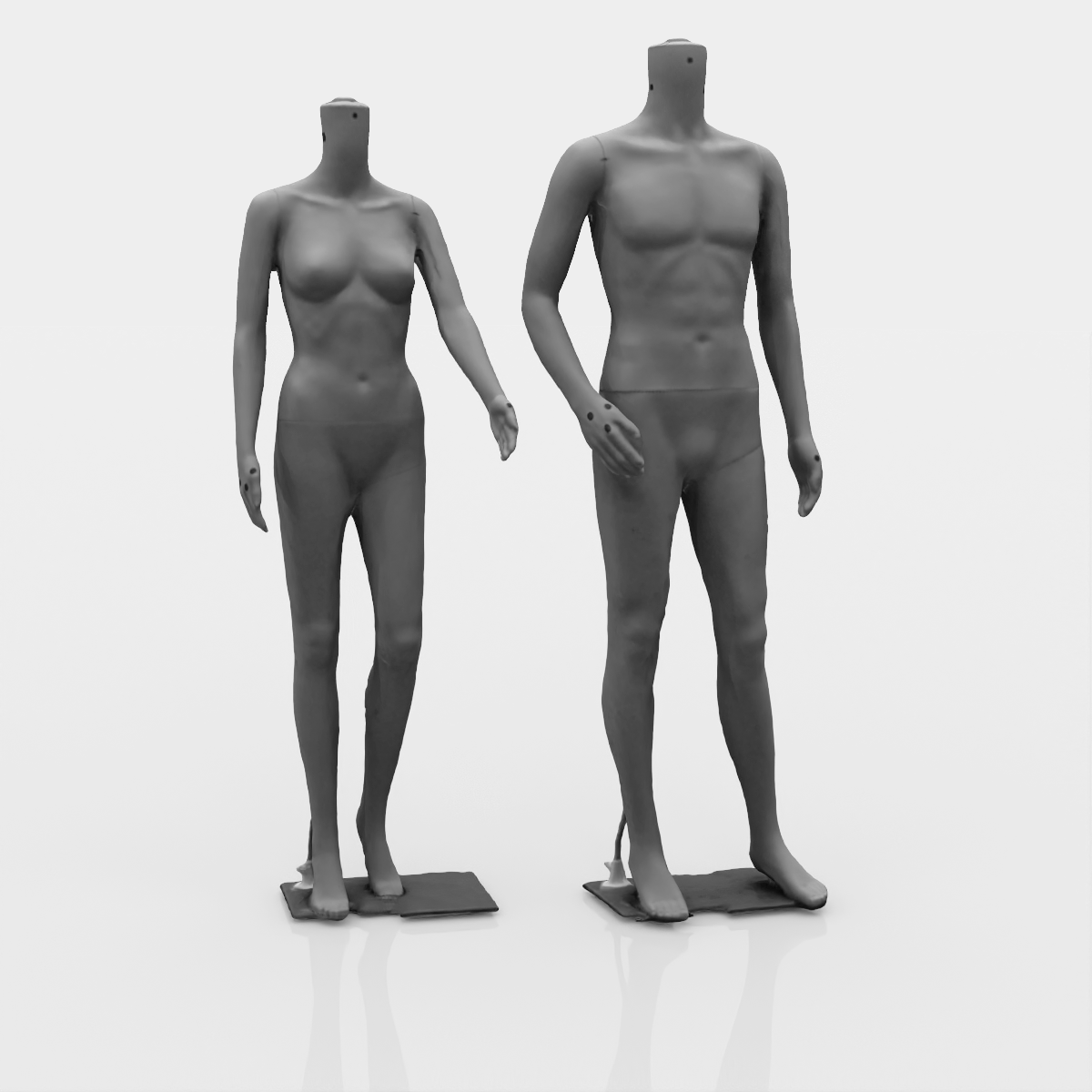}
  \caption{Scans of standard female and male mannequins meshes.}
  \label{fig:mannequins}
\end{figure}

\begin{figure}[h!]
  \centering
  \includegraphics[width=\linewidth]{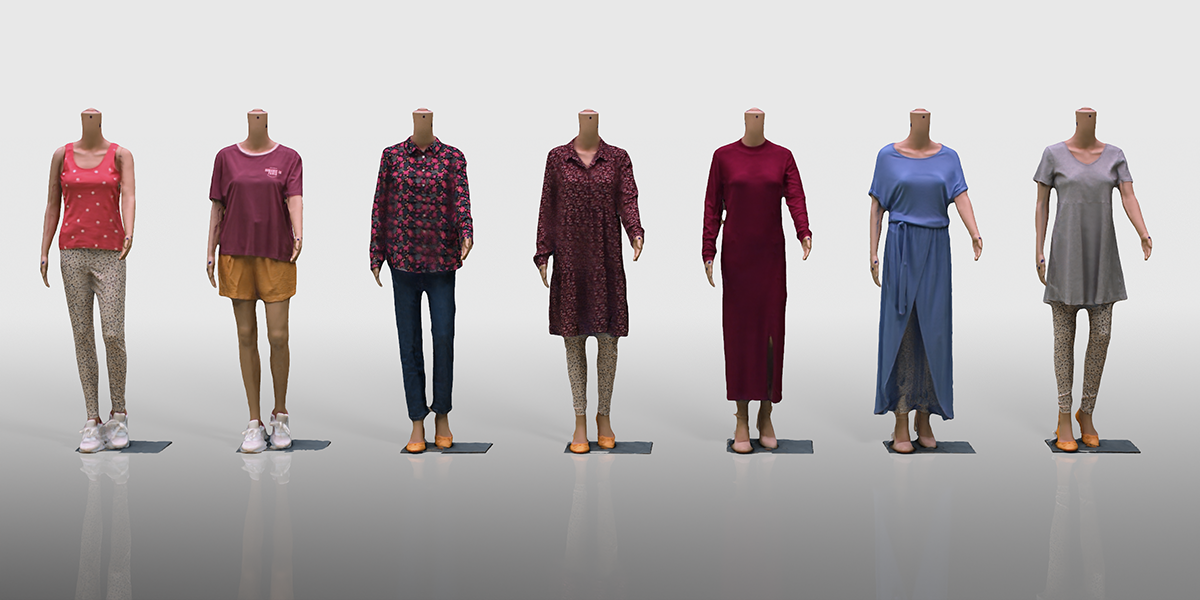}
  \includegraphics[width=\linewidth]{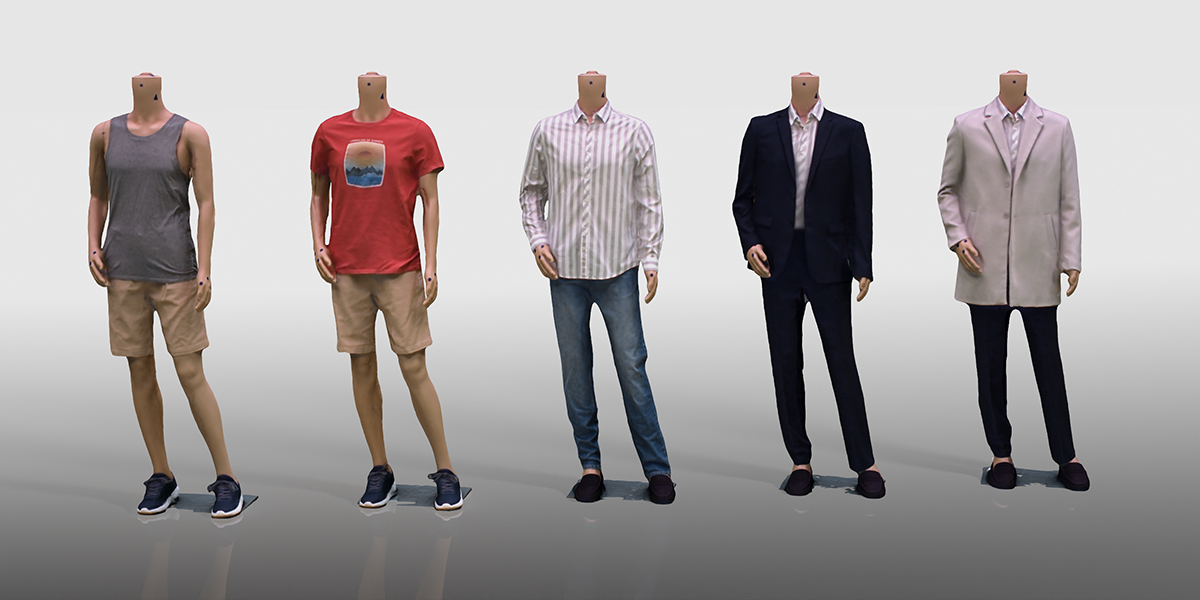}
  \caption{Outfits scanned on female (top) and male (bottom) mannequins.}
  \label{fig:mannequincloth}
\end{figure}

\begin{figure}[h!]
  \centering
  \includegraphics[width=\linewidth]{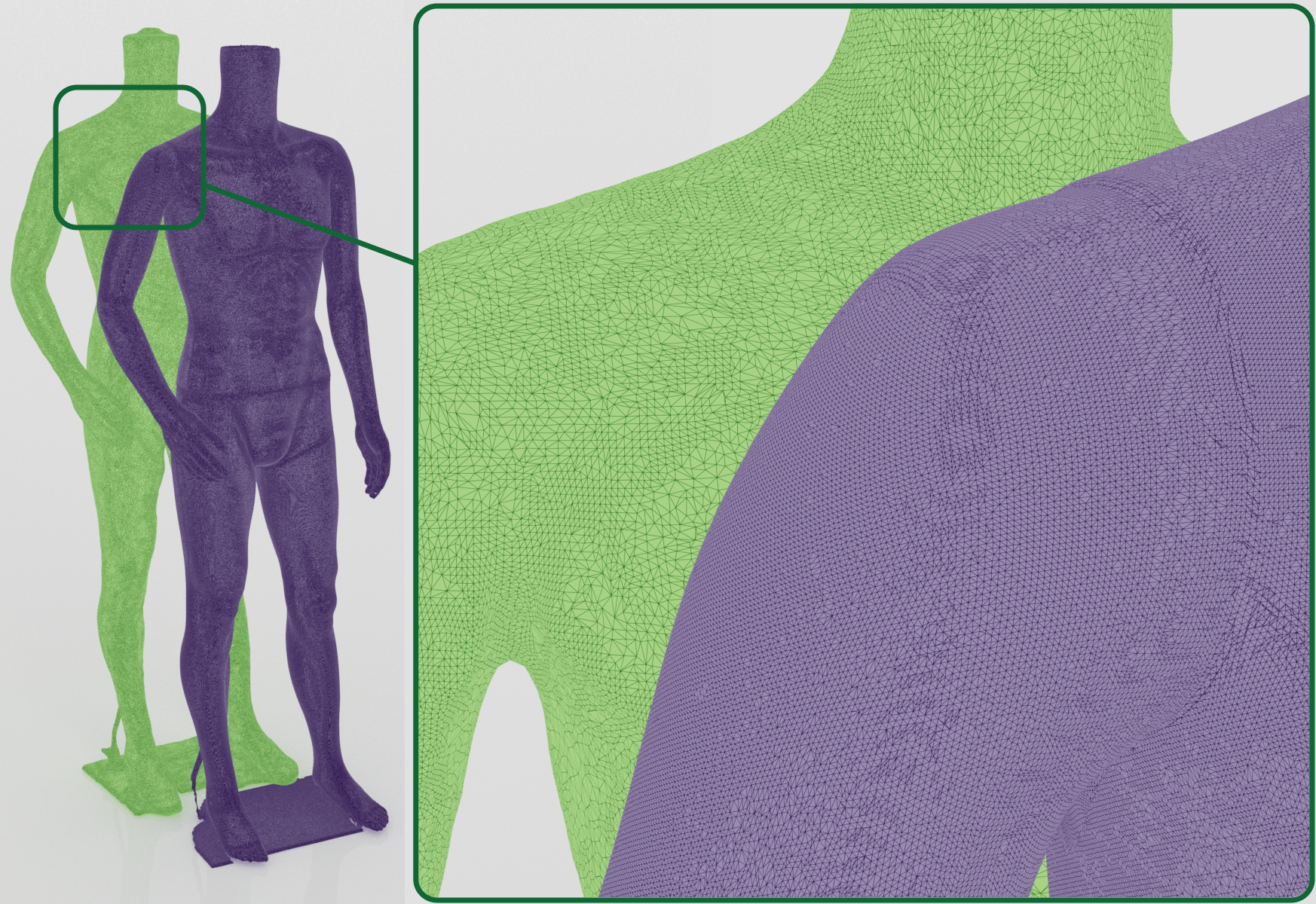}
  \caption{ 250k reconstruction of the male mannequin (green hidden faces rendering) vs. 1500k scanned version (purple hidden faces rendering).}
  \label{fig:rec250kVSscan1500k}
\end{figure}

Fig.~\ref{fig:mannequins} shows the female and male standard mannequins 65k reconstructions without clothing. Fig.~\ref{fig:mannequincloth} shows the same mannequins with all outfits. 
Fig.~\ref{fig:rec250kVSscan1500k} shows the reconstructed male mannequin mesh at resolutions 250k and 1500k.

\subsection{Motions}

\begin{figure}[h!]
  \centering
    \includegraphics[width=\linewidth]{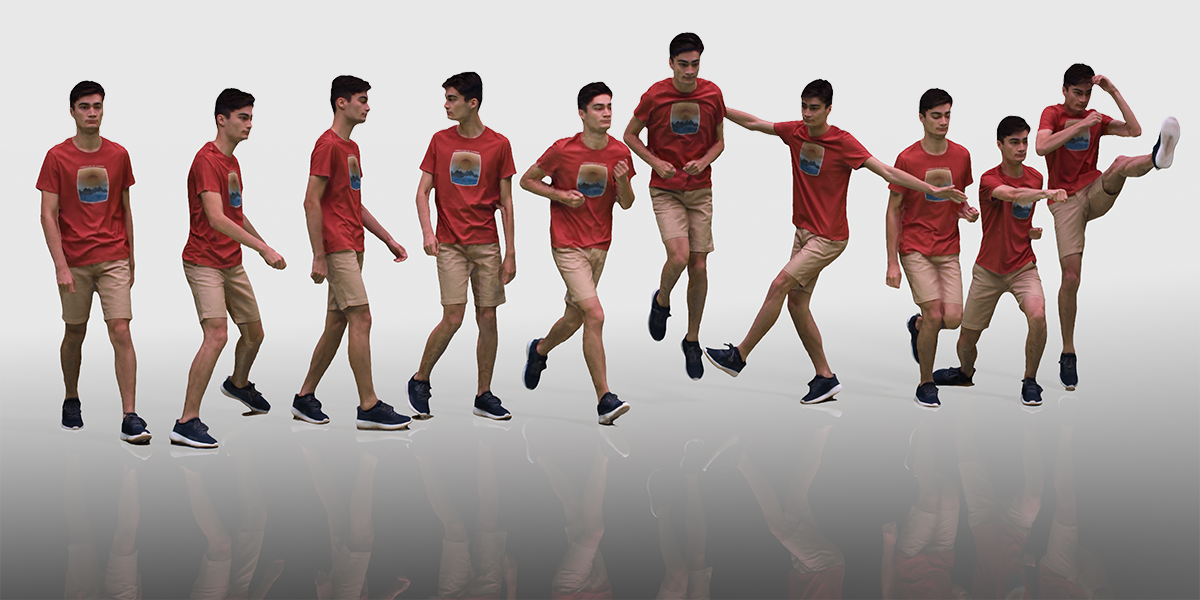}
    \includegraphics[width=\linewidth]{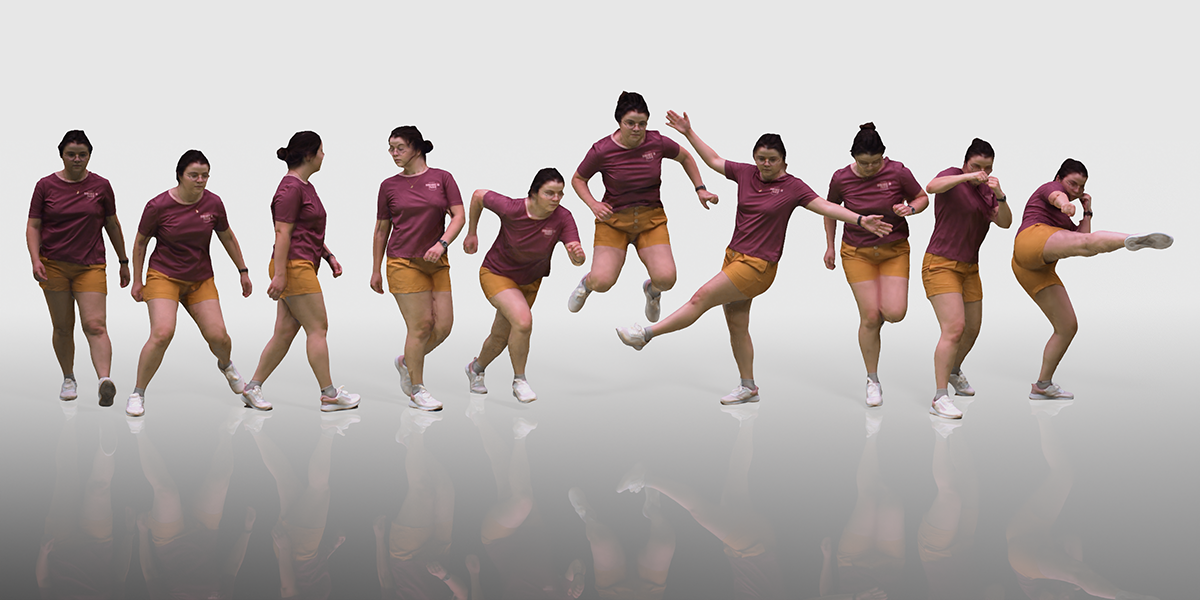}
  \caption{Representative frames of the motions. From left to right: walk, avoid, back, torso, run, jump, dance, hopscotch, free 1, free 2. Here free 1 is boxing, free 2 is kick.  Men (top), women (bottom).}
  \label{fig:motions}
\end{figure}

Actors were asked to perform $11$ motions involving significant displacements within the scene. We focus on motions with large displacements as these are yet rare in existing captured 4D datasets. For instance DYNA~\cite{Dyna:SIGGRAPH:2015} captures soft-tissue dynamics during motions but without large displacements. We further choose the $11$ motions to contain variations in upper and lower body motions, while covering common motions including different variants of walking. The motions, illustrated in Fig.~\ref{fig:motions}, are:
\begin{itemize}
    \item\textbf{walk} a simple walk across the studio;
    \item\textbf{avoid} a walk with last-second obstacle avoidance;
    \item\textbf{back} a walk with a U-turn;
    \item\textbf{torso} a walk with a torso rotation to look backwards;
    \item\textbf{run} a jog / run across the studio; 
    \item\textbf{jump} jump on the spot;
    \item\textbf{dance} a dance with both legs and arms wide motion;
    \item\textbf{hop} hopscotch; 
    \item\textbf{2 free motions} to be chosen by the actor to increase the variability of the dataset, this included mostly martial art, dance and other sport motions;
    \item\textbf{hidden} we recorded one additional motion which will not be released to allow for future evaluations on unobserved data.
\end{itemize}

The duration of the recorded sequences ranges from $0.8$ (for a free motion) to $17.2$ (for dance motion) seconds.

\subsection{Summary}

\begin{figure}[h!]
    \includegraphics[width=\linewidth]{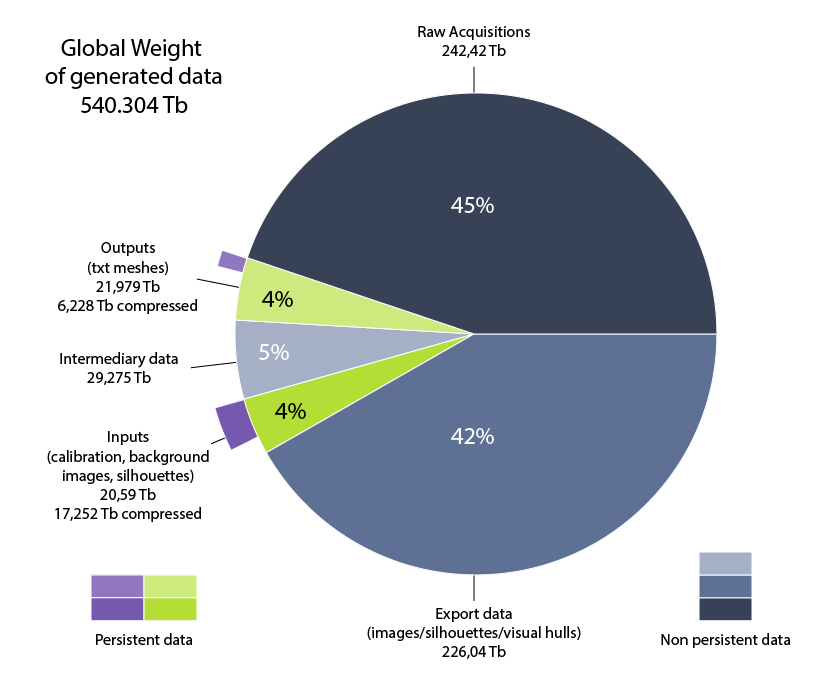}
    \caption{Storage volume required during data processing.}
    \label{fig:weight}
\end{figure}

A total of $1617$ sequences were recorded, involving the processing of $459 080$ frames. The computations were handled on $2$  clusters (17 16-core servers equipped with Nvidia Quadro 4000 cards and 20 16-core Intel Xeon CPU servers) resulting in the generation of 540TB of total data during the project. Fig.~\ref{fig:weight} provides an overview of the storage space required by the data during the generation of 4DHumanOutfit. The final 4D dataset consists of meshes in different resolutions (250k, 65k, 30k, 15k vertices), and undistorted and masked images and silhouettes. The total volume is 22TB, 20.5TB of which are occupied by the undistorted and masked images and silhouettes.

\begin{figure}[h!]
    \includegraphics[width=\linewidth]{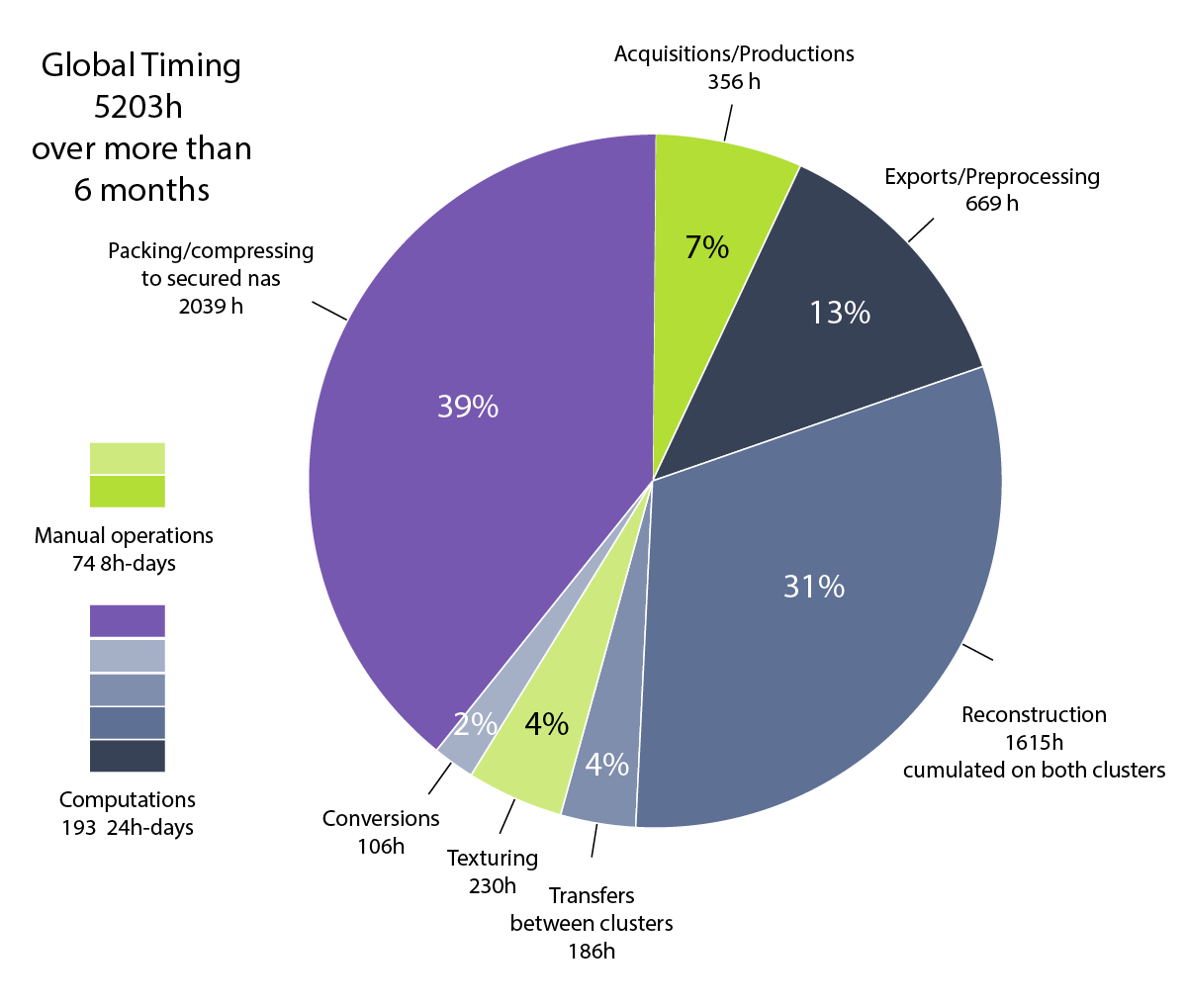}
    \caption{Computation time required during data processing.}
    \label{fig:time}
\end{figure}

Concerning the timing, the project stretched over a bit more than 6 months, for about 5200 hours, divided as shown in Fig.~\ref{fig:time}. A significant amount of time was dedicated to packing and compressing. 

\section{Evaluation}

The 4DHumanOutfit datacube allows to learn correlations between identity, outfit and motion. To demonstrate its potential, we perform a simple evaluation along each of the three factors independently. This demonstrates that each axis of the datacube provides unique information that can be exploited in practical applications. 

\subsection{Identity}
\label{sec:identity}
The first evaluation aims to estimate the body shape of the identity performing a motion from an arbitrary sequence of the datacube. That is, given as input a 4D motion sequence showing a person (of arbitrary identity and outfit) in motion, the aim is to estimate the undressed body shape in T-pose. This problem has been studied previously in  \eg\cite{Yang_2016,Zhang_2017_CVPR}, and is of interest for virtual change room applications.

\paragraph{Evaluation protocol}

To evaluate the accuracy of the retrieved identity in T-pose, we compute for each identity a reference solution of the naked body shape. This is achieved by fitting a parametric human body model to each frame of the person captured in minimal clothing, and by combining the resulting information. That is, we use the minimal clothing regime, which is very close to the actor's skin, as proxy for true body shape.

We use a parametric human body model with two sets of parameters, one representing body shape and one representing static pose. Changing the static pose parameters to the ones representing a T-pose allows to re-pose the body shape into standard pose. In practice, we use the SMPL model~\cite{loper2015smpl}.

To reliably fit the body model to the sequences captured in minimal clothing, we use an existing framework~\cite{MvSMPLfitting} based on SMPLify~\cite{SMPLify}. For a given timestep, we compute 2D keypoints in all images with alphapose~\cite{alphapose}, and optimize a SMPL body model with respect to said keypoints, with an additional loss to force the reprojection of the SMPL mesh to fit inside the silhouette on all images, and the pose and shape priors used in SMPLify.

This is done for all sequences captured in minimal clothing. We then select a fit that minimizes the Chamfer distance to the corresponding 3D reconstruction. The resulting body shape parameters are used to reconstruct a model in T-pose, which is used as reference body shape. For each identity, its reference body shape is released along with the dataset.

To quantitatively evaluate the quality of a body shape estimate computed from a dressed 4D motion sequence of identity $i$, we compute the Chamfer distance between the result and the reference body shape of $i$, in a standard T-pose.

\paragraph{Baseline}

As a baseline method for estimating body-shape parameters, we use the same optimization method described above, based on keypoints and silhouettes, but applied directly to dressed sequences. The setting is more complex, as estimated keypoints are generally less precise on frames with loose clothing, and silhouettes are farther from the silhouette of the body shape. This baseline is computed per-frame.

\paragraph{Results}

We give here numerical and qualitative results of the baseline. The optimization is sensitive to the input keypoints, so results tend to be noisy. When computing the reference shape, this problem is addressed by using the 3D reconstruction to select the best fit. However, the results of the baseline shape estimation are affected by this problem. It sometimes leads to overly small or elongated shapes, when the optimization does not converge to a good solution. In cases of convergence, estimated body shapes are often too big, as the loose clothing is larger than the body shape.

\begin{figure*}[h!]
	\begin{tabular}{c c c c c c c}
		\includegraphics[height=3.2cm]{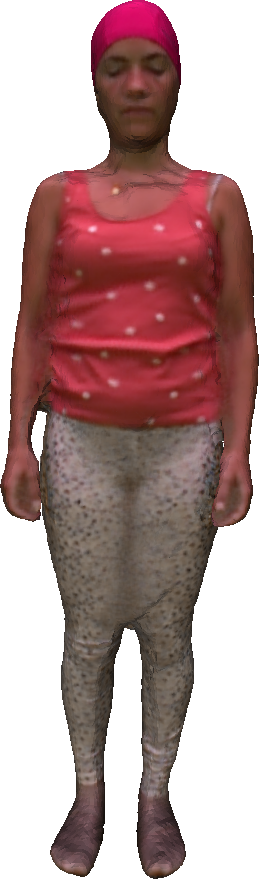} &
        \includegraphics[height=3.2cm]{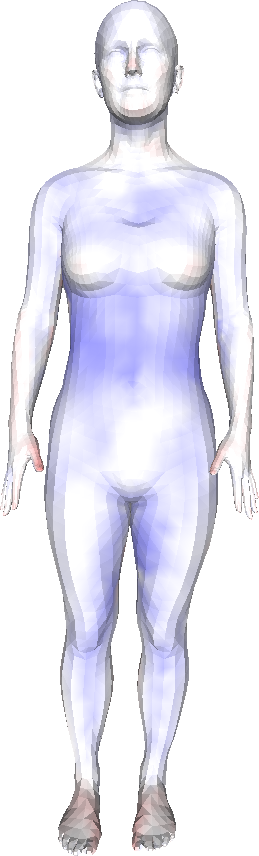} &
        \includegraphics[height=3.2cm]{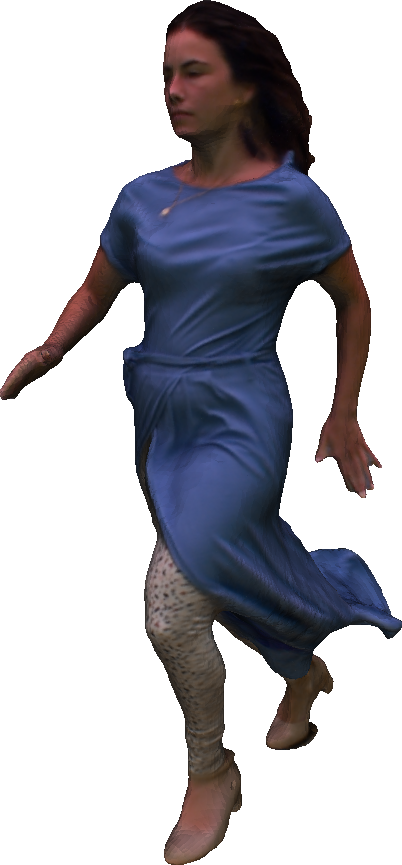} &
        \includegraphics[height=3.2cm]{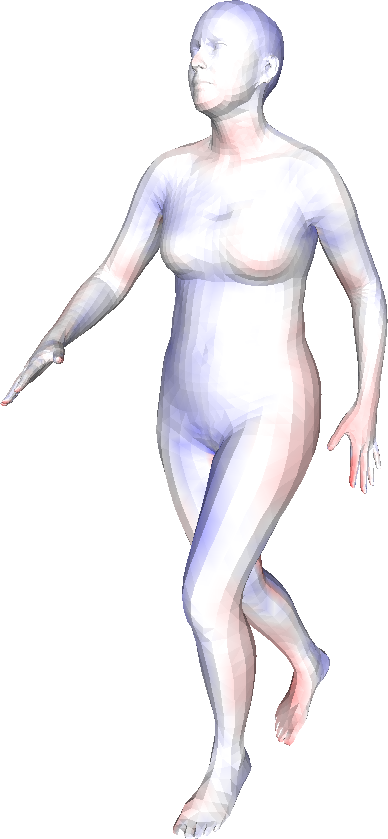} &
        \includegraphics[height=3.2cm]{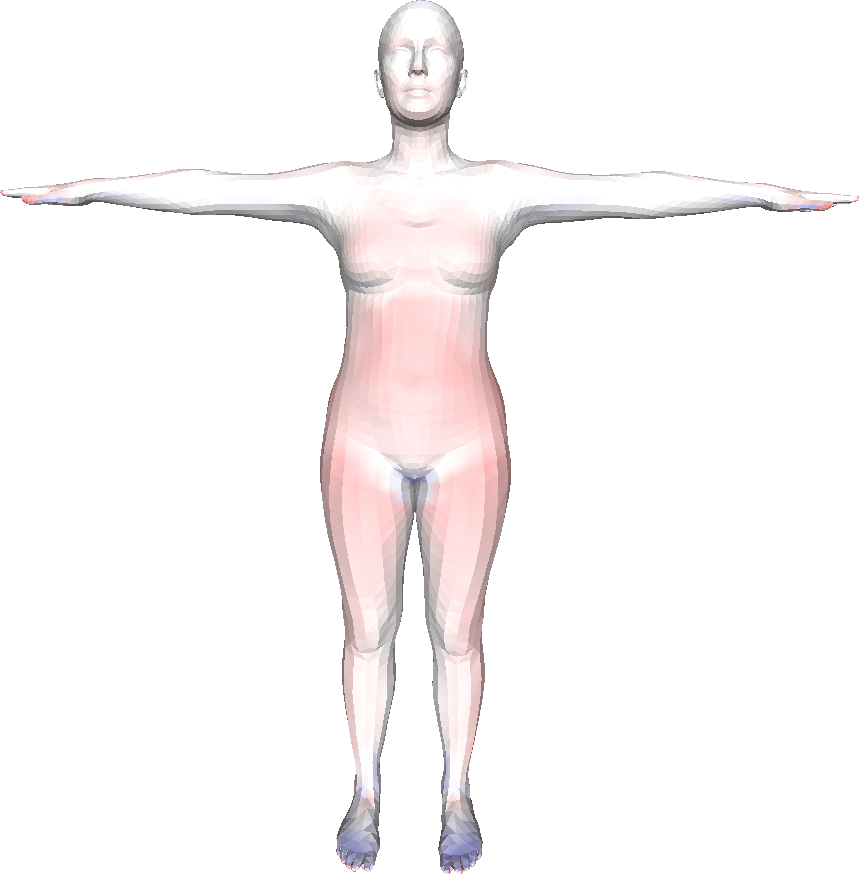} &
        \includegraphics[height=3.2cm]{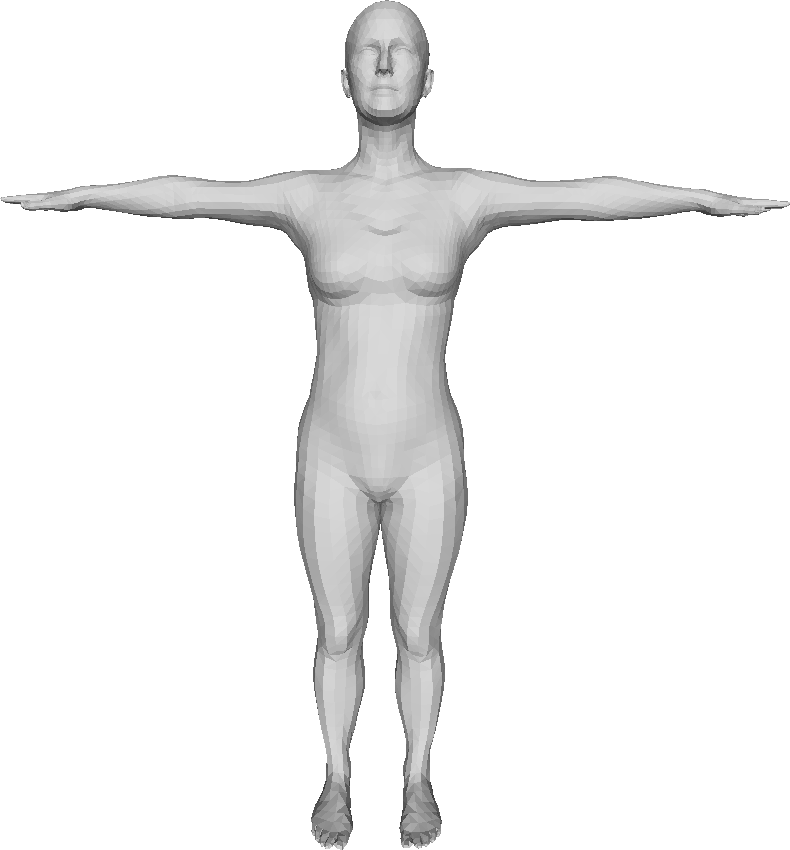} &
        \includegraphics[height=3.2cm]{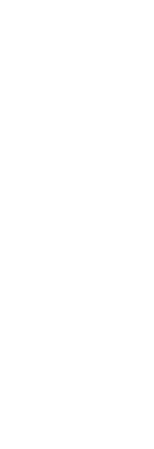} \\
        \includegraphics[height=3.2cm]{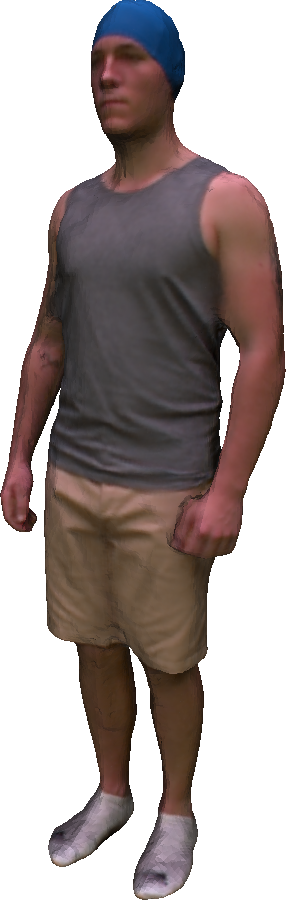} &
        \includegraphics[height=3.2cm]{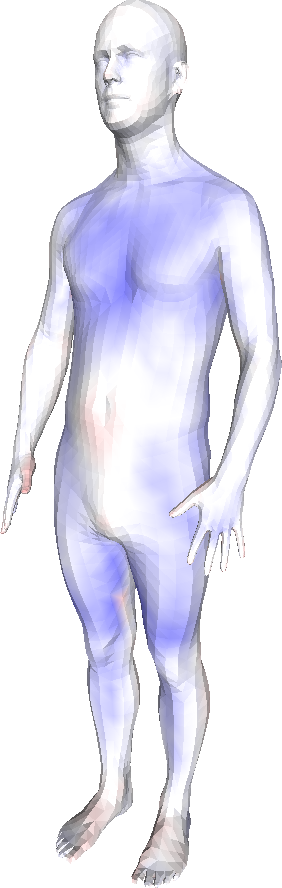} &
        \includegraphics[height=3.2cm]{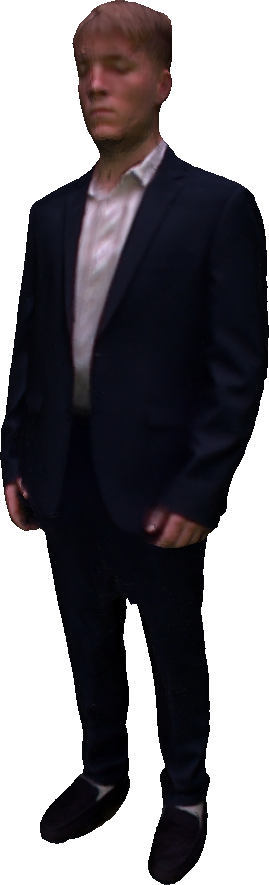} &
        \includegraphics[height=3.2cm]{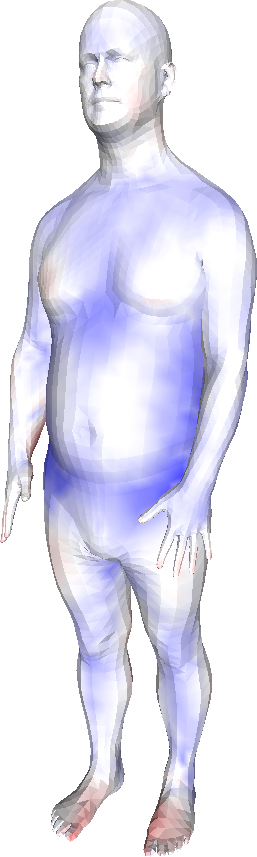} &
        \includegraphics[height=3.2cm]{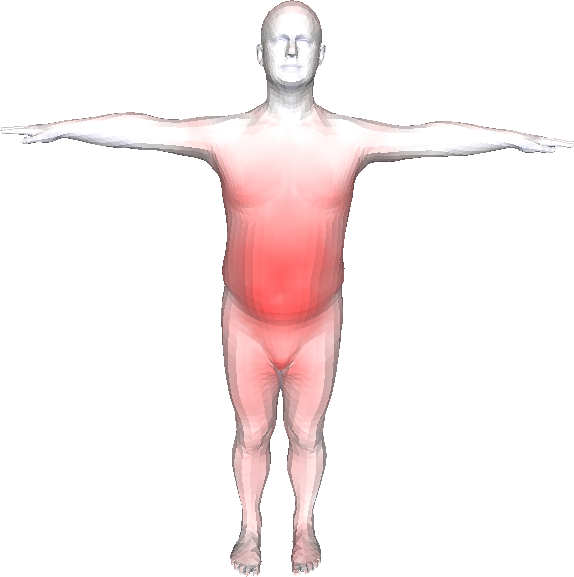} &
        \includegraphics[height=3.2cm]{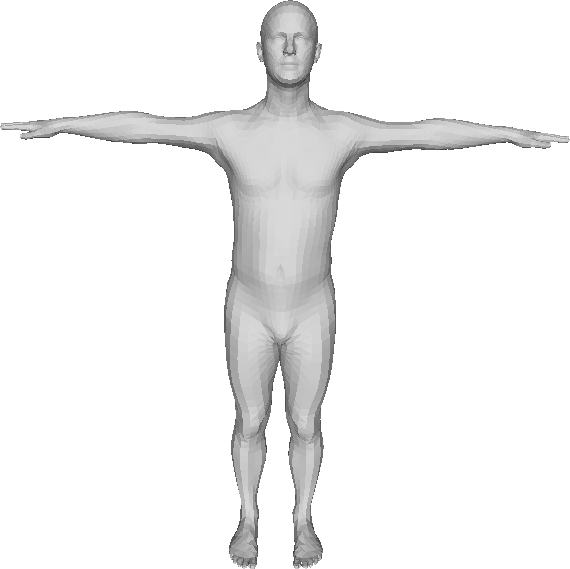} &
        \includegraphics[height=3.2cm]{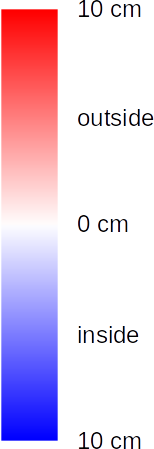} \\
		(a) & (b) & (c) & (d) & (e) & (f) & \\
	\end{tabular}
    \caption{Identity evaluation. Given a sequence in minimal clothing (a), the naked body shape is estimated. This allows to estimate a \textit{reference} body shape for evaluation (b). Vertices are color-coded according to the distance to the reconstructed surface (a). Given an arbitrary clothed sequence as input (c), the naked body shape is estimated (d). Reposing both this result (e) and the reference shape (f) into T-pose allows for quantitative comparison, shown as color-coding.}
    \label{fig:smpl_fit}
\end{figure*}

Fig.~\ref{fig:smpl_fit} shows color-coded results for female \textbf{opt} and male \textbf{cos} outfits. In these examples, the body shape estimates computed by our baseline are close to the reference shape in terms of height and overall body shape. However, the volume of the body shape is overestimated in areas that are occluded by clothing. The reason is that the baseline fits the largest body model that can fit in the silhouettes, and does so on a per-frame basis.

\paragraph{Limitations}
For privacy concerns, we use sequences in tight clothing to compute reference shapes. While this causes small errors due to clothing folds and the width of the cloth, the resulting error is small compared to typical human motions.

We chose a simple baseline to illustrate the challenges of this task. It could be improved by taking information from the full sequence into account.

\subsection{Outfit}

The second evaluation aims to retrieve information related to outfit. Let $J_{\mathcal{M}}^{3d}$ denote the pose of the mannequin shown in Fig.~\ref{fig:mannequins}. Given as input a 4D motion sequence showing a dressed person in motion, the aim is to retrieve the outfit in pose $J_{\mathcal{M}}^{3d}$. This way of evaluating the outfit independently of identity and dynamics is novel to our knowledge. The related problem of inverse cloth design, where the goal is to retrieve a rest pose unaffected by physical forces for use in a physical simulator given the shape of a garment, has been studied~\cite{casati:hal-01309617}, but differs from our scenario as we are interested in retrieving the shape of the outfit in standard pose (as affected by physical forces).

Generating 3D garment deformations using physics-based simulators is challenging, because the generation of realistic detailed 3D garment models is typically done by trained artists and costly, and because physical simulators require input parameters that need to be tuned. 4DHumanOutfit contains accurate 3D garment deformations for a number of outfits, and has the potential to be used for data-driven garment synthesis without relying on physics-based simulators.

\paragraph{Evaluation protocol}
To evaluate the accuracy of the retrieved outfit in pose $J_{\mathcal{M}}^{3d}$, we use the reference scans acquired for each outfit draped on the mannequin as pseudo ground truth. The retrieved outfit is compared to its corresponding reference scan by measuring the Chamfer distance between the two shapes. In particular, we report the Chamfer distance in $mm$ between the retrieved garment mesh $\mathcal{V}_{retr}$ and the reference scan $\mathcal{V}_\mathcal{M}$.

\paragraph{Baseline}
We propose a simple baseline to solve this problem by framing it as a retrieval task. We first compute the 3D joints of each frame in a 4D motion sequence by fitting SMPL to the models as described in the previous section, and then retrieve the frame whose pose $J_{t}^{3d}$ is closest to $J_{\mathcal{M}}^{3d}$. The pose distance is computed by considering a Procrustes-aligned distance as
\begin{equation}
\label{eqn:prcst}
\underset{J_{t}^{3d}}{\argmin}\left(\sum_i \textrm{D}_{P}( J_{t}^{3d}, J_{\mathcal{M}}^{3d})\right),
\end{equation}
where $\textrm{D}_{P}$ is the distance in joint angle space. The 3D joints on the mannequin are obtained by manually annotating 3D points on the surface of the mannequin scan.

\begin{figure}[h!]
  \centering
    \includegraphics[width=\linewidth]{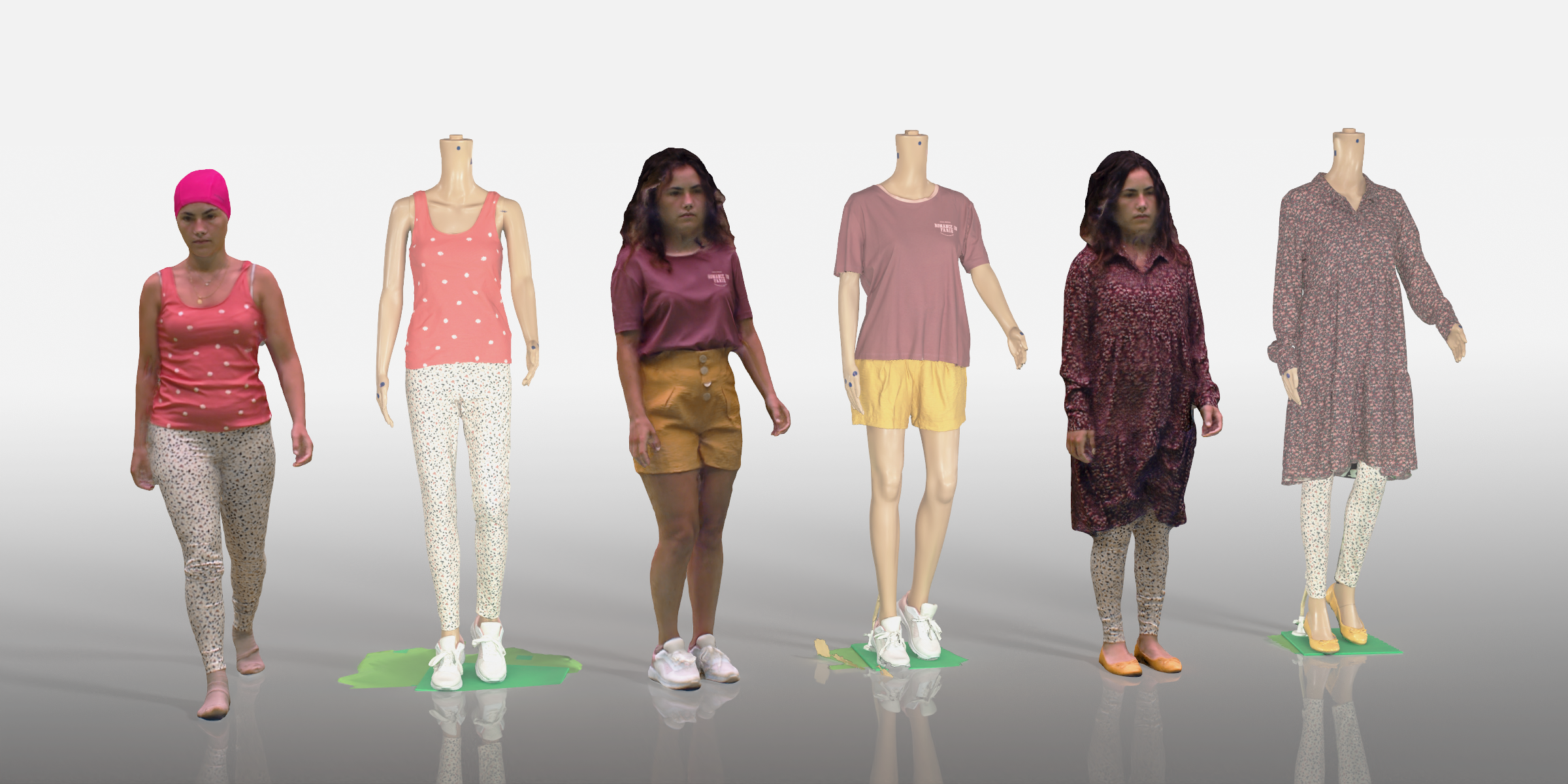}
    \includegraphics[width=\linewidth]{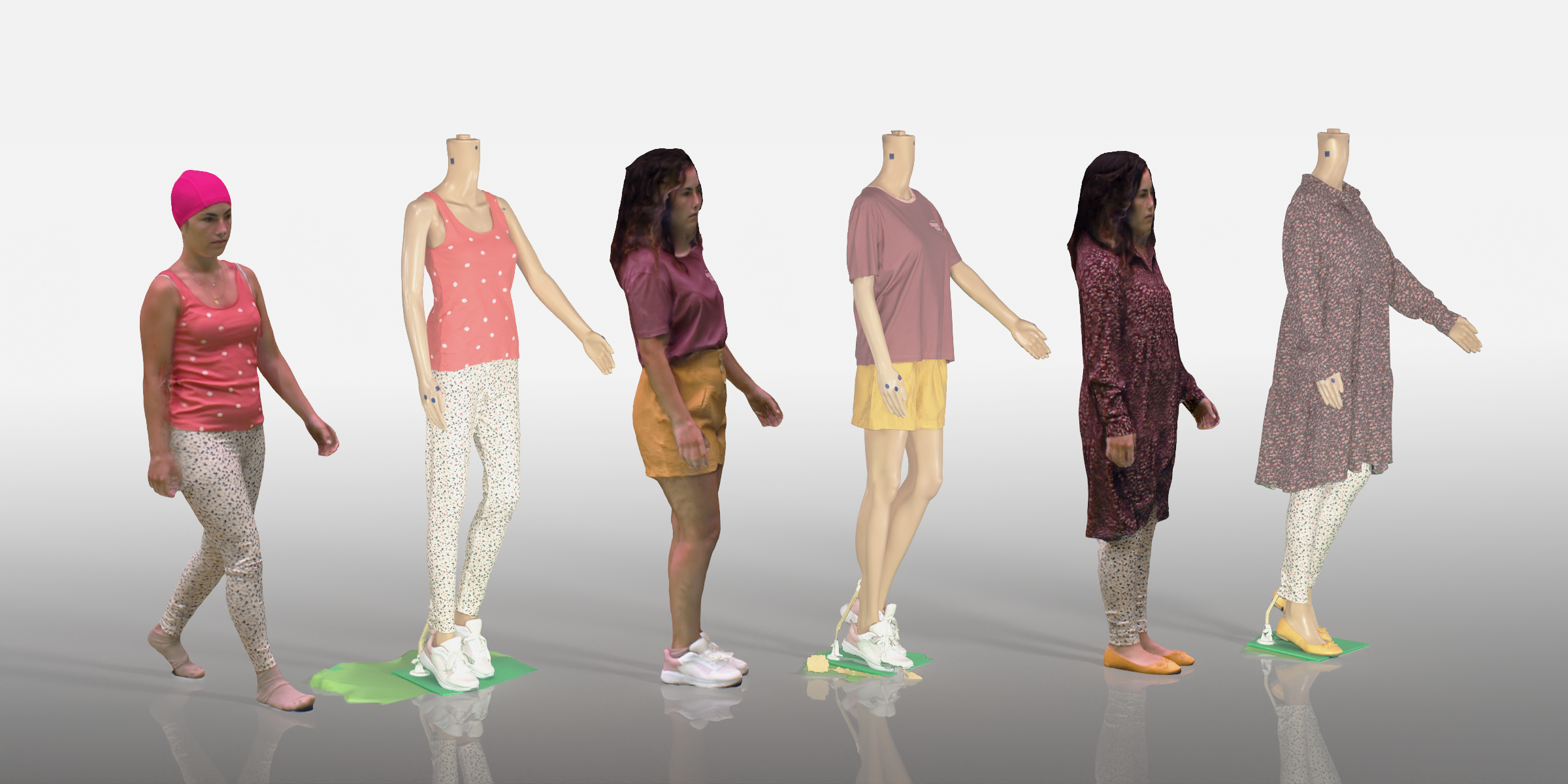}
    \includegraphics[width=\linewidth]{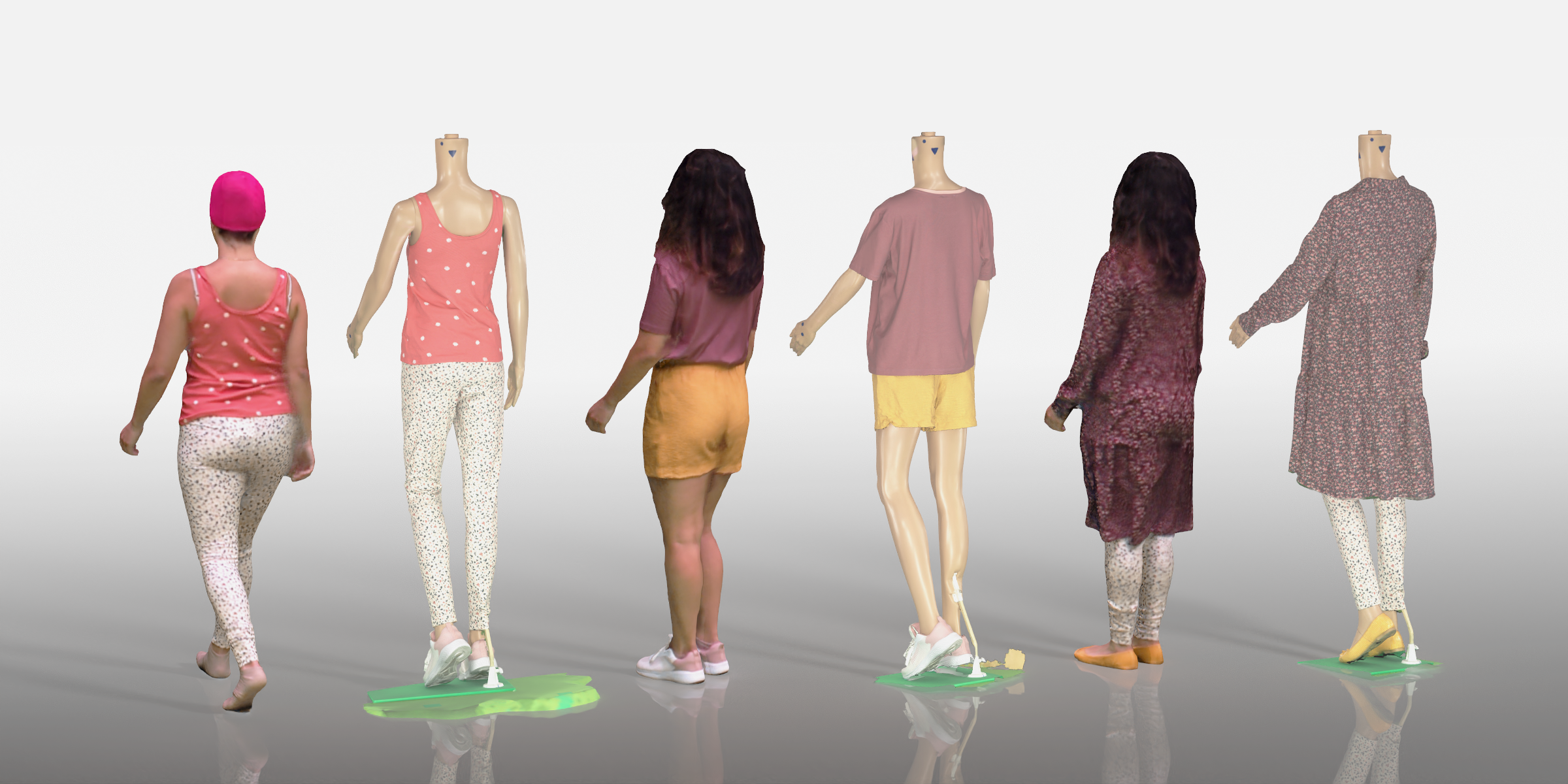}
  \caption{Outfit evaluation. Qualitative visualization of the retrieved outfit. Retrieved outfits and reference scans shown for 3 different outfits (left to right: tig, sho and cos). Three different viewpoints are shown.}
  \label{fig:gar}
\end{figure}

\paragraph{Results}
Results are analyzed for three different subjects wearing three different outfits. Fig.~\ref{fig:gar} shows three viewpoints of the retrieved garment and the reference scan for each of the three subjects. Note that the simple baseline already retrieves poses with garments that have visually similar wrinkle patterns. Tab.~\ref{tab:gar} reports the Procrustes-aligned distance from Eq.~\ref{eqn:prcst}. The error ranges from $26mm$ to $64mm$ across different subjects. This error has high variance because the retrieved outfit is one frame of the input sequence. If the input 4D motion does not contain frames in a pose similar to $J_{\mathcal{M}}^{3d}$, the error is high. Furthermore, the reference scan does not contain head surface, and the mannequin's body shape may not be close to the body shape of the input sequences. 

\paragraph{Limitations}
We propose a novel outfit retrieval task that has potential applications in online garment retrieval. The task along with the reference solutions that allow for quantitative evaluation have the potential to allow for further research in garment retrieval. 

A major limitation of our protocol is that dynamic effects and body shape changes are currently not considered. That is, we frame the problem as a static one even though dynamic motion is present in the 4D motion sequence, and we ignore the influence of the wearer's body shape on the outfit geometry. The baseline we propose is simple and leaves significant room for improvement. However, it already demonstrates that outfit configurations visually similar to the reference scans can be found.

\begin{table}[h]
    \centering
    \scalebox{0.9}{
    \begin{tabular}{|c|c|c|}
    \hline
          &$\mathcal{V}_{retr} \rightarrow \mathcal{V}_\mathcal{M}$ &   $\mathcal{V}_\mathcal{M} \rightarrow \mathcal{V}_{retr}$ \\\hline
          tig & 63.8 & 60.1 \\\hline
          sho & 34.0 & 30.6 \\\hline
          cos & 26.0 & 29.8 \\\hline
    \end{tabular}
    }
    \caption{Quantitative outfit evaluation. Chamfer distance (in $mm$) between retrieved garment mesh $\mathcal{V}_{retr}$ and $\mathcal{V}_\mathcal{M}$ for outfits \textbf{tig}, \textbf{sho} and \textbf{cos}.}
    \label{tab:gar}
\end{table}

\subsection{Motion}

The third evaluation aims to examine the motion axis. Our evaluation considers the task of motion retargeting where the objective is to generate a 4D motion sequence of a given identity that performs the same motion as another given 4D motion sequence. In particular, given as input the 4D motion sequence showing identity $i_1$ in outfit $o$ performing motion $m$ along with a 3D model of identity $i_2$ wearing minimal clothing, the goal is to compute the 4D sequence showing identity $i_2$ while performing motion $m$.  

A challenge when evaluating motion retargeting is the lack of realistic ground truth data. On the one hand, realistic 4D ground truth is lacking due to the sparse nature of existing large 4D human datasets~\eg\cite{mahmood2019amass,cai2022humman} where all actors are not seen performing all motions. This lack of data encourages state of the art to evaluate on synthetically generated 3D motions. These synthetic motions are often generated with skinning methods and lack realistic local dynamic details. On the other hand, smaller 4D datasets~\cite{dfaust:CVPR:2017,Dyna:SIGGRAPH:2015} with dynamic details do not account for clothing.  

In the following, we show that 4DHumanOutfit can be leveraged to evaluate motion retargeting methods by providing captured reference solutions with accurate geometric details.  

\paragraph{Evaluation protocol}

To evaluate the accuracy of the retargeted motion, we compare the 4D motion resulting from the retargeting to the target motion of identity $i_2$ in minimal clothing performing motion $m$ captured in 4DHumanOutfit. 

In this scenario, the retargeted motion $M_1 = \{m_{1,j}\}_{j=1}^{n}$ and the target motion $M_2 = \{m_{2,j}\}_{j=1}^{m}$ are characterized by sequences of 3D point clouds. The point clouds are not in correspondence, so we use the Chamfer distance to compare them. To compare $M_1$ and $M_2$, the Chamfer distance relies on a nearest neighbor search per point, which is heavily influenced by small variations of the global trajectory and temporal unfolding of $M_1$ and $M_2$. We remove this influence by spatio-temporally aligning $M_1$ and $M_2$, as this is common when evaluating retargeting approaches~\cite{marsot2023correspondencefree,jiang2022h4d}. 

To align the global trajectories, we center the pointclouds using their centroid $c$. To align the temporal unfolding of the motions, we use Dynamic Time Warping (DTW)~\cite{dtw1994}. 
Given two sequences of point clouds, DTW computes the optimal monotonic path $p^*$ between aligned frames as 
{\small
\begin{equation}
    \nonumber
    p^* = \underset{p}{\argmin}\left(\sum_j \textrm{D}_{Ch}( m_{1,j}-c_{1,j}, m_{2,p[j]}-c_{2,p[j]})\right),
\end{equation}
}
where $\textrm{D}_{Ch}$ is the Chamfer distance. The proposed metric is then evaluated as the median error along this path as 
{\small
\begin{equation}
    \underset{j}{med}\left( \textrm{D}_{Ch}\left( m_{1,j}-c_{1,j}, m_{2,p^*[j]}-c_{2,p^*[j]}\right) \right).
    \label{eq:retarget}
\end{equation}
}

\paragraph{Baseline}

Motion retargeting has been approached from different angles, either using deformation models that directly operate on the body surface~\cite{wang2020neural,regateiro2022temporal}, using structured latent representations of 4D motion~\cite{jiang2022h4d,marsot22} or using skeletal representations which are linked to the body surface using an animation model~\cite{villegas21,marsot2023correspondencefree}. 

Some of these works require correspondences between the target and source bodies or temporal correspondences in the source motion. Our source motion is unregistered and we can leverage the template fitting from Sec.~\ref{sec:identity} to have access to a target identity in minimal clothing in T-pose. From the applicable methods~\cite{marsot2023correspondencefree,wang2020neural,jiang2022h4d}, we choose~\cite{marsot2023correspondencefree} as our baseline because it generalizes well under various motion and shape preservation metrics and was already tested on the raw data of a multi-view acquisition setup. 
The baseline operates in three steps. First, the source skeleton is extracted using a PointFormer network. Second, this skeletal motion is retargeted to the target at the skeletal level using a recurrent network. Third, the dense geometry of the target shape is recovered using a learnt skinning prior. 

\paragraph{Results}

\begin{figure}
    \centering
    \includegraphics[width=\columnwidth]{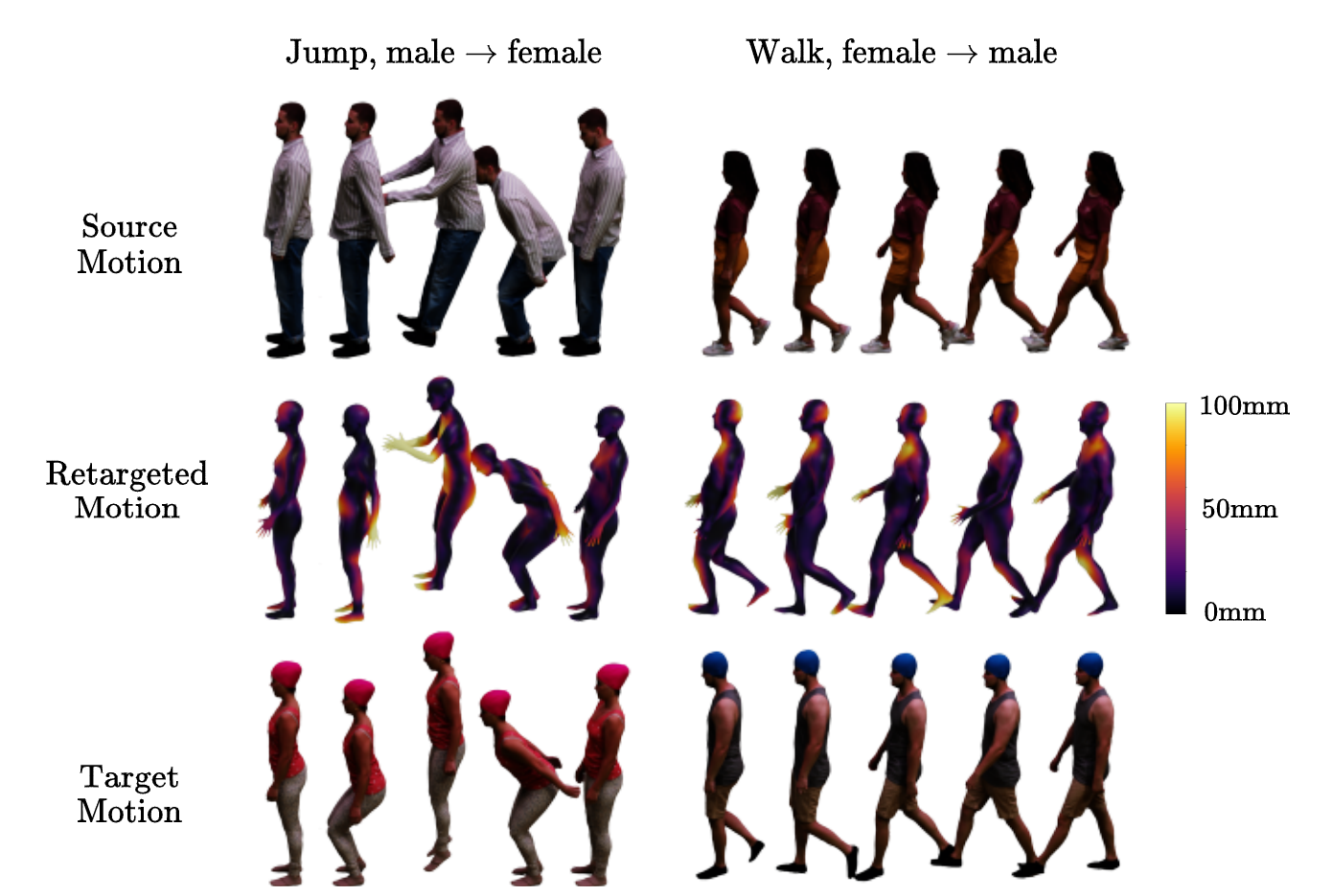}
    \caption{Motion evaluation. Qualitative visualization of the Chamfer distance for two retargetings. Top: Source motion with outfit. Middle: Retargeted motion, color coded by Chamfer distance to the target motion. Bottom: Target motion in tight clothing.}
    
    \label{fig:moret}
\end{figure}

Tab.~\ref{tab:moret} reports the metric of Eq.~\ref{eq:retarget} for 4 retargetings considering 2 identities: a female and a male identity for 2 source motions and 2 source outfits. As the per sequence median error is
more informative when comparing different methods, we also visualize the spatio-temporal distribution of the error for two retargetings to differentiate error due to the natural variability and the error introduced by the retargeting method in Fig.~\ref{fig:moret}. 

Fig.~\ref{fig:moret} shows the retargeting from male to female on a jumping motion and from female to male on a walking motion. The color coding shows that the method generates plausible poses overall with large error (yellow color) due to natural variability in the arm pose between the source and target  jumping motions. It also highlights that the method could be improved in terms of head and arm pose transfer (red color) with the head facing down in the jumping retargeting and incorrect arm poses in some frames of the walking retargeting. 
\begin{table}[h]
    \centering
    \scalebox{0.9}{
    \begin{tabular}{|c|c|c|}
    \hline
          &male,jea $\rightarrow$ female &   female ,sho $\rightarrow$ male \\\hline
          walk & 0.020 & 0.026 \\\hline
          jump & 0.026 & 0.035 \\\hline
    \end{tabular}
    }
    \caption{Motion evaluation. Median Chamfer distance (Eq.~\ref{eq:retarget} in $m$) for 4 retargeting examples.}
    \label{tab:moret}
\end{table}

\paragraph{Limitations}

It is known that a fixed type of motion performed by the same performer is performed slightly differently at different trials~\cite{Ghorbani2020}. This variability is not implemented in our baseline, which instead outputs a deterministic retargeting solution. To address this limitation, our evaluation protocol  normalizes global trajectory and temporal unfolding.

Second, existing retargeting baselines that operate on raw scan data are limited to outfits that are close to the body surface. Hence, we cannot leverage more ample outfits present in 4DHumanOutfit. 

\section{Conclusion}
We presented 4DHumanOutfit, a large-scale dataset of 20 actors wearing 7 outfits each and performing 11 motions per outfit. This data captures detailed spatio-temporal dynamics of varying outfits, and their interaction with different morphologies and motions. We demonstrated that each axis of the resulting data-cube contains unique information using simple evaluations. This data has the potential of serving in many different applications involving digital humans including augmented or virtual reality applications (\eg virtual change rooms), and in entertainment (\eg animation content generation).

\section{Acknowledgments}
This work was supported by French government funding managed by the National Research Agency under grants ANR-21-ESRE-0030 (CONTINUUM), ANR-19-CE23-0013 (3DMOVE), and ANR-19-CE23-0020 (Human4D).

{\small
\bibliographystyle{abbrv}
\bibliography{references}
}

\end{document}